%% file: main.tex
\documentclass[10pt,twocolumn,letterpaper]{article}

\usepackage[pagenumbers]{cvpr} 

\input{preamble}
\definecolor{cvprblue}{rgb}{0.21,0.49,0.74}
\usepackage[pagebackref,breaklinks,colorlinks,allcolors=cvprblue]{hyperref}
\usepackage[absolute,overlay]{textpos}
\usepackage{wrapfig} 
\usepackage[table]{xcolor} 
\usepackage{booktabs}
\usepackage{graphicx}
\usepackage{booktabs}   
\usepackage{graphicx}   
\usepackage{makecell}   
\usepackage{pifont}
\usepackage{tabularx}
\usepackage{cuted}
\usepackage{multirow} 
\renewcommand\theadfont{\small\bfseries}
\def\paperID{*****} 
\def\confName{CVPR}
\def\confYear{2026}

\title{VideoSSR: Video Self-Supervised Reinforcement Learning}

\author{
    Zefeng He\textsuperscript{1,2}
    \quad Xiaoye Qu\textsuperscript{1}\textsuperscript{*}
    \quad Yafu Li\textsuperscript{3}
    \quad Siyuan Huang\textsuperscript{4}
    \quad Daizong Liu\textsuperscript{5}
    \quad Yu Cheng\textsuperscript{3}\thanks{Corresponding authors} \\
    \textsuperscript{1}Shanghai Artificial Intelligence Laboratory, 
    \textsuperscript{2}Nanjing Univerisity \\ 
    \textsuperscript{3}The Chinese University of Hong Kong \\
    \textsuperscript{4}Shanghai Jiao Tong University, 
    \textsuperscript{5}Wuhan University
}

\begin{document}
\maketitle
\input{sec/0_abstract}    
\input{sec/1_intro}

\input{sec/2_related}

\input{sec/3_method}

\input{sec/4_experiment}

\input{sec/5_conclusion}

{
    \small
    \bibliographystyle{ieeenat_fullname}
    \bibliography{main}
}

\input{sec/X_suppl}

\end{document}

%% file: sec/0_abstract.tex
\begin{abstract}
Reinforcement Learning with Verifiable Rewards (RLVR) has substantially advanced the video understanding capabilities of Multimodal Large Language Models (MLLMs). However, the rapid progress of MLLMs is outpacing the complexity of existing video datasets, while the manual annotation of new, high-quality data remains prohibitively expensive.
This work investigates a pivotal question: Can the rich, intrinsic information within videos be harnessed to self-generate high-quality, verifiable training data?
To investigate this, we introduce three self-supervised pretext tasks: Anomaly Grounding, Object Counting, and Temporal Jigsaw. 
We construct the Video Intrinsic Understanding Benchmark (VIUBench) to validate their difficulty, revealing that current state-of-the-art MLLMs struggle significantly on these tasks. Building upon these pretext tasks, we develop the VideoSSR-30K dataset and propose VideoSSR, a novel video self-supervised reinforcement learning framework for RLVR. Extensive experiments across 17 benchmarks, spanning four major video domains (General Video QA, Long Video QA, Temporal Grounding, and Complex Reasoning), demonstrate that VideoSSR consistently enhances model performance, yielding an average improvement of over 5\%. These results establish VideoSSR as a potent foundational framework for developing more advanced video understanding in MLLMs.
The code is available at \url{https://github.com/lcqysl/VideoSSR}.

\end{abstract}



%% file: sec/1_intro.tex
\section{Introduction}
\label{sec:intro}

\begin{figure}[h!]
    \centering
    \includegraphics[width=\columnwidth]{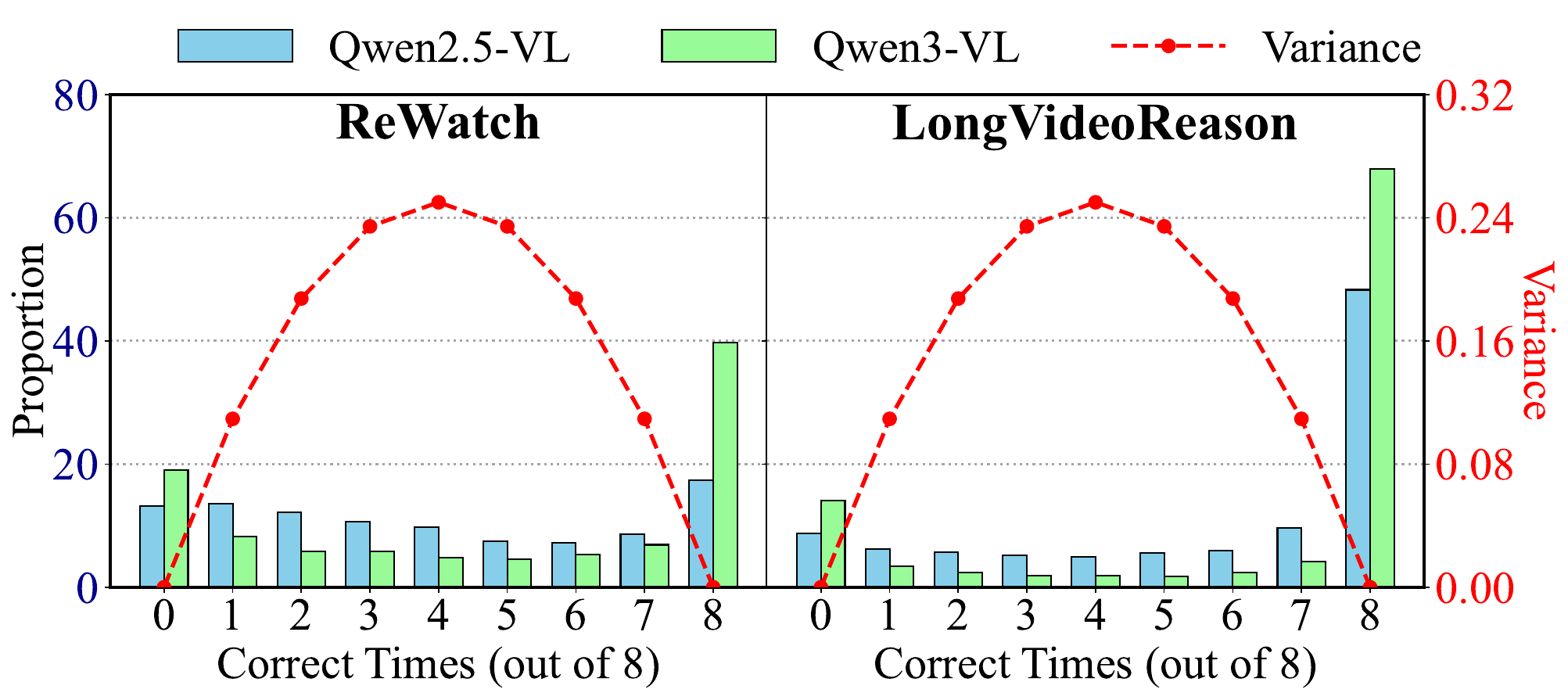}
    
    \caption{\textbf{Distribution of answer correctness on ReWatch and LongVideoReason.} 
    Across both models and datasets, a vast majority of questions yield a bimodal outcome, 
    resulting in either zero or eight correct answers. This zero variance issue is notably 
    more pronounced for the more powerful Qwen3-VL model.}
    \vspace{-10pt}
    \label{fig:difficulty} 
\end{figure}

In past years, Multimodal Large Language Models (MLLMs) have achieved remarkable progress in the field of video understanding~\cite{Qwen2.5VL,gpt4o,Gemini2.5,gemini1.5,OpenAI2025-GPT5,bai2025intern,wang2025internvl3,qwen3vl}. 
Benefiting from recent Reinforcement Learning with Verifiable Reward (RLVR)~\cite{deepseekmath,deepseekr1,feng2025videor1,li2025videochatr1,rewatch-r1,tang2025video,zhang2025survey}, the performance of MLLMs has been further improved. 
A cornerstone of the RLVR approach is the availability of video datasets with verifiable answers. To obtain the verifiable answers, existing datasets, such as LongVideoReason~\cite{chen2025longvila-r1} and ReWatch~\cite{rewatch-r1}, utilize multi-agent collaboration to construct high-quality datasets with verifiable answers. 

\begin{figure*}[t!]
    \centering
    \includegraphics[width=\textwidth]{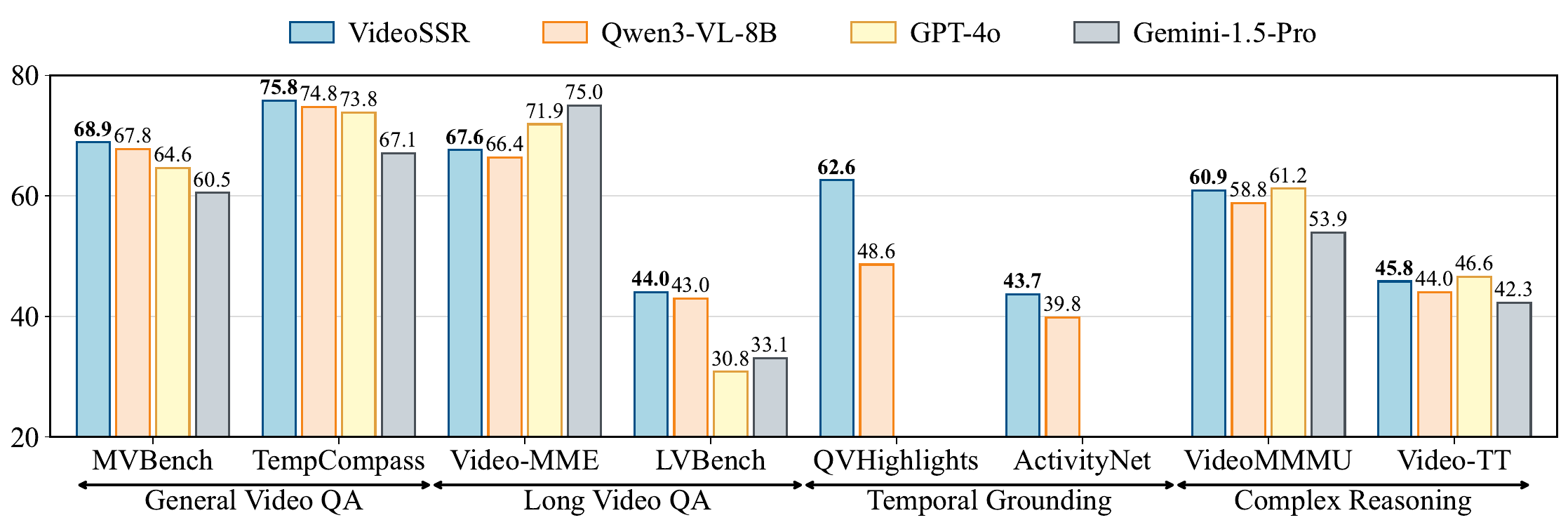}
    \caption{\textbf{Performance comparison on four video tasks}. Input frames for VideoSSR and Qwen3-VL-8B do not exceed 64.}
    \label{fig:overall_framework}
\end{figure*}

Although current datasets have effectively enhanced the performance of models like Qwen2.5-VL~\cite{Qwen2.5VL}, significant limitations arise when applying them to more powerful models, such as the recent Qwen3-VL~\cite{qwen3vl}. 
First, for highly capable models, many questions in existing datasets lack sufficient complexity to serve as effective training challenges. 
To illustrate this, we generate eight independent responses per question using Qwen2.5-VL~\cite{Qwen2.5VL} and Qwen3-VL~\cite{qwen3vl}. As shown in Figure~\ref{fig:difficulty}, a vast majority of questions yield a perfect score where all eight responses are correct, indicating they are insufficiently challenging.
Second, the multi-agent annotation process introduces systemic biases and artifacts, which create flawed or spurious reward signals for RLVR, particularly when the annotator models are less capable than the target models. 
This is evidenced by another large portion of questions where all generated responses are incorrect, suggesting either intractable difficulty or biased ground truths. The resulting bimodal distribution of scores, with most questions exhibiting zero variance, offers an ineffective learning signal for GRPO~\cite{deepseekmath,deepseekr1} training in RLVR. Consequently, training advanced models on such data yields marginal gains or even performance degradation (Section~\ref{sec:ablation_study}).


Compounding these issues is the prohibitive cost of manual annotation for video. This predicament, however, points to a compelling alternative: can the rich, intrinsic information within videos be harnessed to construct high-quality, verifiable questions for RLVR? 
Inspired by traditional video self-supervised learning~\cite{noroozi2017representation,noroozi2016unsupervised,xu2019self,fernando2017self}, we first design three self-supervised pretext tasks with parametrically scalable difficulty, including Anomaly Grounding, Object Counting, and Temporal Jigsaw, to generate verifiable questions.
To validate the difficulty of these tasks, we construct \textbf{V}ideo \textbf{I}ntrinsic \textbf{U}nderstanding \textbf{Bench} (\textbf{VIUBench}) and found that questions targeting the intrinsic properties of the video itself remain profoundly challenging, 
even for leading closed-source models like 
GPT-5~\cite{OpenAI2025-GPT5}.

Building on this insight, we introduce \textbf{VideoSSR}, a new \textbf{Video }\textbf{S}elf-\textbf{S}upervised \textbf{R}einforcement learning framework to enhance the video understanding of MLLM.
We construct the VideoSSR-30K dataset using the aforementioned pretext tasks, which is 
entirely independent of human or MLLM annotations. 
This dataset is subsequently 
utilized to train our model with GRPO.
To overcome the challenge of sparse reward signals arising from the inherent difficulty of these tasks, we design corresponding smooth reward functions for each pretext task to ensure efficient and stable RLVR training.

To validate the generalization capability of VideoSSR, we conduct extensive experiments on 17 benchmarks spanning four main video tasks: General Video QA, Long Video QA, Temporal Grounding, and Complex Reasoning. 
The results show that our proposed VideoSSR achieves consistent performance improvements across all benchmarks and under three different input frame settings, demonstrating an average gain of over 5\%. 

In summary, our main contributions are fourfold:
\begin{itemize}

    \item We generate verifiable training data for RLVR that harnesses intrinsic video signals. This self-supervised paradigm circumvents the prohibitive costs and inherent biases of prevailing multi-agent and manual annotation, thereby addressing a critical bottleneck in scaling MLLMs for video understanding.

    \item We introduce three self-supervised pretext tasks with parametrically scalable difficulty. Moreover, we construct VIUBench benchmark from these tasks, which reveals profound limitations in state-of-the-art MLLMs for intrinsic video understanding.

    \item We introduce VideoSSR, a self-supervised reinforcement learning framework for video RLVR training and construct VideoSSR-30K dataset. To facilitate efficient and stable RLVR training, in VideoSSR, we further design three tailored smooth reward functions.

    \item Extensive experiments across 17 benchmarks demonstrate the superior generalization capability of VideoSSR, consistently achieving average performance improvements exceeding 5\% and establishing it as a foundational approach for advancing video understanding.
    


    
\end{itemize}

%% file: sec/2_related.tex
\begin{figure*}[t]
    \centering
    \includegraphics[width=\textwidth]{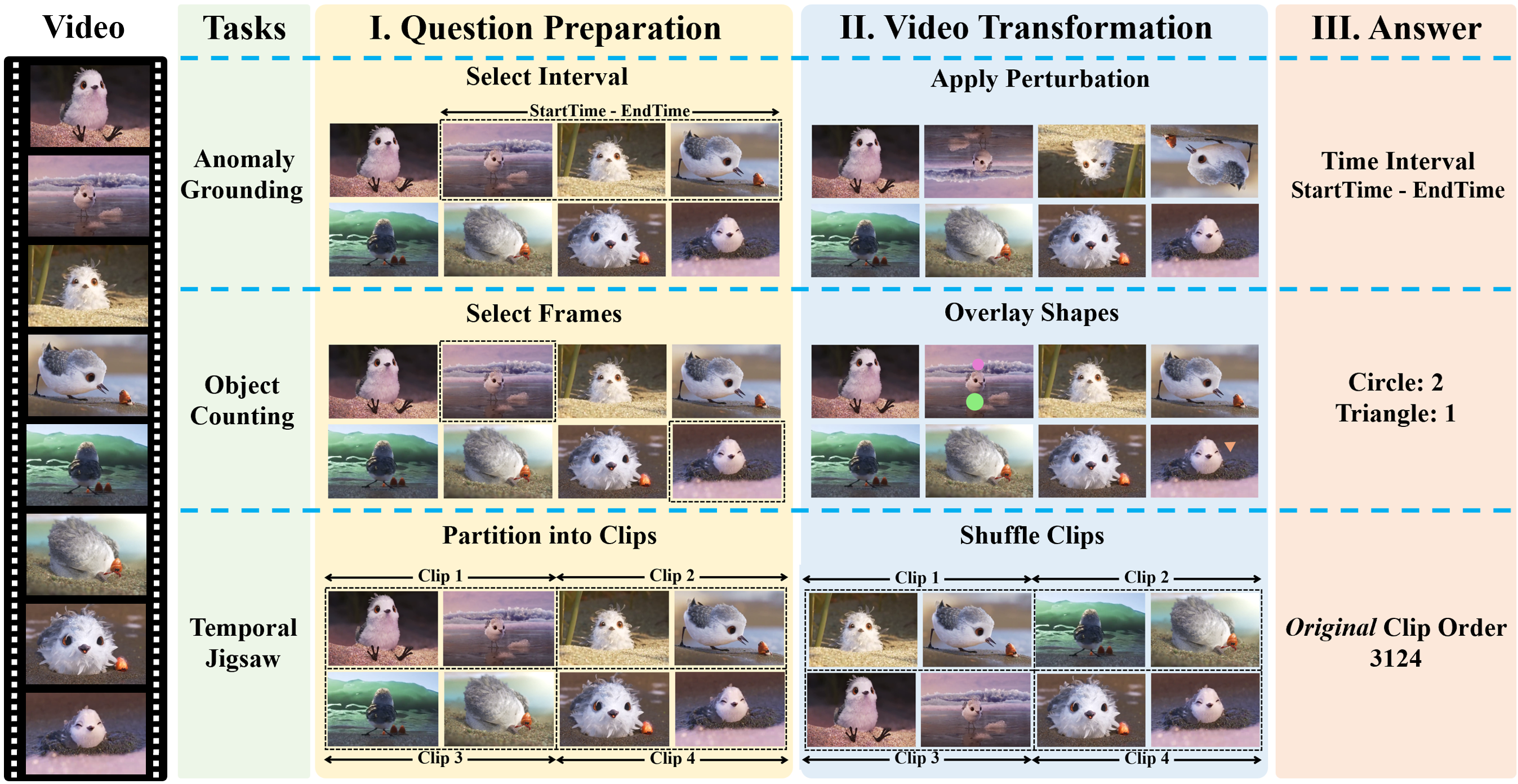}
    \caption{\textbf{An overview of our three self-supervised pretext tasks.} 
    \textbf{(a) Anomaly Grounding:} A temporal segment is perturbed (e.g., via rotation), and the task is to identify the start and end timestamps of this anomaly. 
    \textbf{(b) Object Counting:} Procedurally generated shapes are overlaid onto selected frames, and the task is to count the total number of each shape type. 
    \textbf{(c) Temporal Jigsaw:} The video is divided into clips which are then shuffled. The task is to predict the original temporal order of the segments.}
    \label{fig:pretext_tasks}
\end{figure*}

\section{Related Works}
\label{sec:formatting}

\subsection{Reinforcement Learning for MLLMs}
Reinforcement Learning with Verifiable Reward (RLVR)~\cite{deepseekmath,deepseekr1} has been shown to significantly enhance the reasoning capabilities of language models, a success that has been rapidly extended to MLLMs~\cite{chen2025grpo,chen2025longvila-r1,li2025videochatr1,he2025framethinker,yan2025videochat,dong2025videotg,qu2020fine}. 
For instance, Video-R1~\cite{feng2025videor1} leverages existing Video QA 
datasets~\cite{zhang2024video,xiao2021next,wu2024star,patraucean2023perception,yi2019clevrer} 
to bolster performance on Video QA tasks. Time-R1~\cite{wang2025time} utilizes datasets with precise timestamp annotations to improve Temporal Grounding. 
SpaceR~\cite{ouyang2025spacer} automatically generates verifiable questions from the geometric and semantic ground truths of 3D scenes~\cite{brown2025simsv,yang2025cambrian}, enhancing the model's spatial reasoning abilities. 
ReWatch-R1~\cite{rewatch-r1} leverages multi-agent collaboration to construct high-quality reasoning datasets, thereby advancing its capabilities in complex reasoning.
Despite these diverse data sourcing strategies, several fundamental limitations persist. The reliance on external annotations often introduces significant bias. 
Meanwhile, many approaches often specialize in enhancing a single capability, which can limit their broader generalization.

\subsection{Self-supervised learning for Video}

Self-supervised learning~\cite{noroozi2017representation,noroozi2016unsupervised,xu2019self,fernando2017self,wang2020self,jing2020self,schiappa2023self,liu2022unsupervised,liu2024unsupervised} for video aims to learn effective spatio-temporal representations from unlabeled video data. The core principle involves designing pretext tasks that capitalize on the inherent properties of video. For instance, early works leverage tasks such as video jigsaw puzzles~\cite{noroozi2016unsupervised,xu2019self,wang2022video,huo2021self,ahsan2019video} to learn representations. Similarly, recent research~\cite{zeng2025agentic,wu2025visual} has employed the jigsaw puzzle task to facilitate the reinforcement learning of MLLMs. 
While training MLLMs with the video jigsaw task has been shown to enhance performance on tasks requiring temporal-centric understanding~\cite{wu2025visual}, the self-supervised paradigm has not been fully explored for video understanding. 
In this work, we move beyond a single task and investigate a richer suite of pretext tasks to cultivate more comprehensive generalization in MLLMs.

%% file: sec/3_method.tex
\definecolor{separatorcolorclose}{HTML}{e5f4e5} 
\definecolor{separatorcolor}{HTML}{fef8e6} 
\definecolor{ourscolor}{HTML}{dff2f7}      
\definecolor{grey}{gray}{0.9}

\section{Method}
Considering there is rich information in the video, in this paper, we explore leveraging the intrinsic information within the video itself to construct high-quality questions with scalable difficulty. To investigate it, we begin by designing novel pretext tasks.

\subsection{Pretext Tasks}
\label{sec:pretext_tasks}

In this section, we introduce three pretext tasks, including Anomaly Grounding, Object Counting, and Temporal Jigsaw. 
These tasks share a common design philosophy, namely, they can generate verifiable question-answer pairs directly from raw videos, independent of any human or model-generated annotations. 
Furthermore, the difficulty of these pairs can be parametrically controlled. The overall process for these three tasks is illustrated in Figure~\ref{fig:pretext_tasks}.

\subsubsection{Anomaly Grounding}
This task assesses the model's ability to localize temporal segments that violate natural video dynamics. 
Let a video be represented as a sequence of frames $V = \{f_1, f_2, \dots, f_T\}$, with a total duration of $D$ seconds. 
We first randomly select a temporal interval $[t_s, t_e] \subseteq [0, D]$, where $t_s$ and $t_e$ are the start and end timestamps, respectively. 
This interval corresponds to a contiguous segment of frames $S = \{f_i \mid \text{timestamp}(f_i) \in [t_s, t_e]\}$.

Next, we apply a perturbation function $\mathcal{P}$ to this segment to create a perturbed version, $S' = \mathcal{P}(S)$. 
The function $\mathcal{P}$ is sampled from a set of predefined transformations targeting different core capabilities:
\begin{itemize}
    \item \textbf{Fine-grained Perception}: e.g., swapping the red and blue color channels for every frame in $S$.
    \item \textbf{Spatial Perception}: e.g., rotating every frame in $S$ by 180 degrees.
    \item \textbf{Temporal Perception}: e.g., randomly shuffling the frame order within $S$.
\end{itemize}

The final video $V'$ is constructed by replacing the original segment $S$ with its perturbed counterpart $S'$. 
The model is then provided with the modified video $V'$ and is tasked to identify the anomalous interval by predicting its start and end timestamps, $(t_s, t_e)$. 

\subsubsection{Object Counting}
This task targets the model's fine-grained perception and counting abilities. 
We define a set of primitive geometric shapes $\mathcal{C} = \{c_1, c_2, \dots, c_K\}$, such as circles, rectangles, and triangles, which can be procedurally generated. 
For a given video $V$, we randomly select a subset of frames $\mathcal{F}_{sub} \subset V$. 
For each frame $f_i \in \mathcal{F}_{sub}$, we synthesize a set of objects $O_i$, where each object $o \in O_i$ is an instance of a shape class from $\mathcal{C}$ with randomized attributes (size, color, rotation, position). 
These modified frames, denoted as $f'_i$, are then used to create the final video $V'$ by replacing their original counterparts.
The ground truth is a vector of counts $\mathbf{n} = [N_1, N_2, \dots, N_K]$, where each element $N_k$ is the total number of occurrences of shape $c_k$:
\begin{equation}
    N_k = \sum_{f_i \in \mathcal{F}_{sub}} \left| \{ o \in O_i \mid \text{type}(o) = c_k \} \right|
\end{equation}
Given $V'$, the model is required to output the counts for each shape category.

\subsubsection{Temporal Jigsaw}
This task is designed to evaluate the model's temporal perception, specifically its understanding of temporal coherence and event ordering.
We partition the video $V$ into $n$ contiguous, non-overlapping segments of equal duration, $V = [S_1, S_2, \dots, S_n]$. 
We then generate a random permutation $\pi$ of the indices $\{1, 2, \dots, n\}$. A new video $V'$ is created by reordering the segments according to this permutation:
\begin{equation}
    V' = [S_{\pi(1)}, S_{\pi(2)}, \dots, S_{\pi(n)}]
\end{equation}
The model is presented with the shuffled video $V'$ and is tasked with restoring the original temporal order. To achieve this, it must predict a sequence of indices that correctly reorders the shuffled segments. This target sequence is the inverse of the permutation $\pi$ that was used for shuffling. The answer is therefore the sequence defined by $\pi^{-1}$:
\begin{equation}
    \text{Answer} = (\pi^{-1}(1), \pi^{-1}(2), \dots, \pi^{-1}(n))
\end{equation}

\subsection{Video Intrinsic Understanding Benchmark}

\begin{figure}[h!]
    \centering
    \includegraphics[width=\columnwidth]{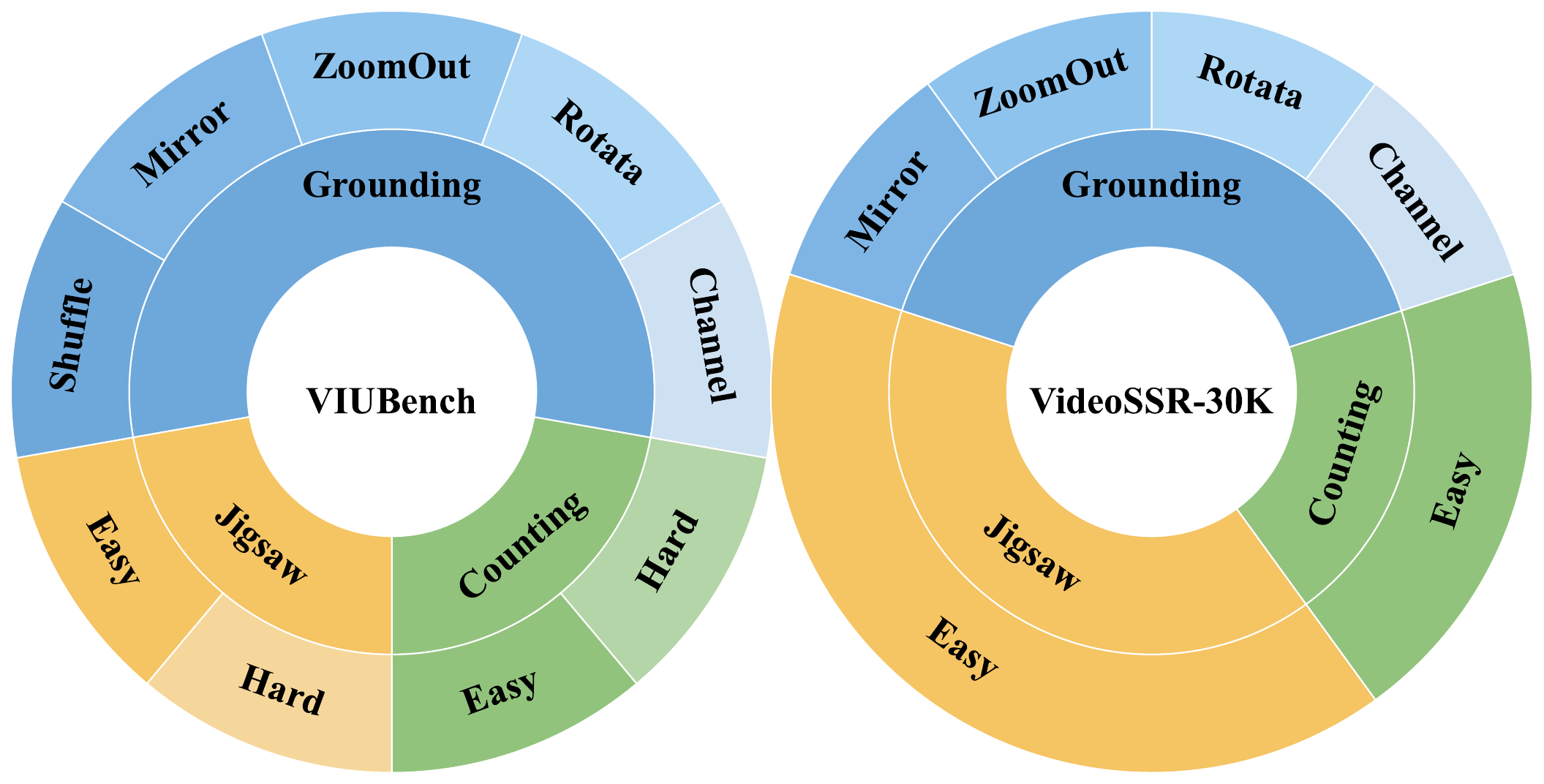}
    
    \caption{\textbf{Task distribution in VIUBench and VideoSSR-30K.} The left panel illustrates the proportional data distribution across our three pretext tasks and their subtypes for VIUBench. The right panel shows the corresponding composition of VideoSSR-30K.}
    
    \label{fig:data} 
\end{figure}

After defining the above three tasks, a critical question arises: are these pretext tasks sufficiently challenging for state-of-the-art MLLMs? 
To investigate this, we construct the \textbf{V}ideo \textbf{I}ntrinsic \textbf{U}nderstanding \textbf{Bench} (\textbf{VIUBench}), which systematically evaluates a model's ability to comprehend intrinsic video properties across three core axes: Fine-grained Perception, Spatial Perception, and Temporal Perception. 
The benchmark is composed of 2700 question-answer pairs generated from our three pretext tasks. The proportional distribution of data across these tasks is illustrated in the left panel of Figure~\ref{fig:data}.

 \label{sec:viubench}

\colorlet{group1color}{Red!10}       
\colorlet{group2color}{Yellow!25}    
\colorlet{group3color}{Blue!10}      

\begin{table*}[t]
\centering
\caption{\textbf{Performance comparison on VIUBench.} The benchmark assesses three core 
abilities (Fine-Grained Perception, Spatial Perception, and Temporal Perception) via our three pretext tasks (Object Counting, Anomaly Grounding, and Temporal Jigsaw). For both open source and closed source models, the top result is shown in bold, and the second-best is underlined.}
\label{tab:viubench}
\resizebox{\textwidth}{!}{%
\begin{tabular}{l cc ccccc cc cccc}
\toprule
{ \textbf{Ability}} & \multicolumn{3}{c}{ \textbf{Fine-Grained Perception}} & \multicolumn{3}{c}{ \textbf{Spatial Perception}} & \multicolumn{3}{c}{ \textbf{Temporal Perception}} & \multicolumn{4}{c}{ \textbf{Average}} \\ 
\cmidrule(lr){2-4} \cmidrule(lr){5-7} \cmidrule(lr){8-10} \cmidrule(lr){11-14} 

{ \textbf{Task}} 
& \multicolumn{2}{c}{\cellcolor{group1color}{ Object Counting}} 
& \multicolumn{5}{c}{\cellcolor{group2color}{ Anomaly Grounding}} 
& \multicolumn{2}{c}{\cellcolor{group3color}{ Temporal Jigsaw}}
& \cellcolor{group1color}{ Counting} 
& \cellcolor{group2color}{ Grounding} 
& \cellcolor{group3color}{ Jigsaw} 
& { Overall} \\

\cmidrule(lr){2-3} \cmidrule(lr){4-8} \cmidrule(lr){9-10}
\cmidrule(lr){11-14}

{ \textbf{Type}} 
& \cellcolor{group1color}{ Easy} & \cellcolor{group1color}{ Hard} 
& \cellcolor{group2color}{ Channel} & \cellcolor{group2color}{ Rotate} & \cellcolor{group2color}{ ZoomOut} & \cellcolor{group2color}{ Mirror} & \cellcolor{group2color}{ Shuffle} 
& \cellcolor{group3color}{ Easy} & \cellcolor{group3color}{ Hard} 
&-- &-- &-- &--\\

\midrule
\multicolumn{14}{c}{\cellcolor{grey}\textit{Random Guess}} \\
\midrule
Random Guess& 11.1 & 6.3 & 25.9 & 25.4 & 25.4 & 25.2 & 25.2 & 0.1 & 0.0 & 8.7 & 25.4 &0.1 & 16.1 \\
\midrule
\multicolumn{14}{c}{\cellcolor{separatorcolorclose}\textit{Closed Source Models}} \\
\midrule
 GPT-5~\cite{OpenAI2025-GPT5} & \textbf{88.4} & \textbf{70.3} & \underline{82.6} & \underline{81.8} & \textbf{56.5} & \underline{48.9} & \underline{34.1} & \textbf{39.0} & \textbf{27.0} & \textbf{79.4} & \underline{60.8} &\textbf{33.0} & \textbf{58.7} \\
Gemini-2.5-Pro~\cite{Gemini2.5} & \underline{80.8} & \underline{61.3} &\textbf{84.6} & \textbf{82.3} & \underline{51.0} & \textbf{55.7} & \textbf{52.1} & \underline{25.3} & \underline{17.7} & \underline{71.1} & \textbf{65.1} &\underline{21.5} & \underline{56.7} \\
 Gemini-2.5-Flash~\cite{Gemini2.5} & 35.7 & 22.4 & 75.8 & 73.7 & 28.5 & 30.2 & 28.6 & 8.3 & 4.0 & 29.1 & 47.4 &6.2 & 34.1 \\
 Seed1.5-VL~\cite{guo2025seed1} & 72.3 & 52.0 & 79.0 & 70.7 & 19.4 & 31.4 & 24.1 & 20.7 & 9.3 &62.6 & 44.9 &15.0 & 42.2 \\
\midrule
\multicolumn{14}{c}{\cellcolor{separatorcolor}\textit{Open Source Models}} \\
\midrule
 Qwen2.5-VL-7B-Instruct~\cite{Qwen2.5VL} & 11.3 & 5.3 & 6.4 & 13.7 & 8.1 & 5.2 & 7.0 & 0.7 & 0.0 & 8.3 & 8.1 & 0.3 & 6.4 \\
 VideoJigsaw-7B~\cite{wu2025visual} & 12.1 & 5.4 & 1.5 & 4.5 & 1.2 & 1.1 & 2.0 & \underline{20.3} & \underline{5.0} & 8.8 & 2.1 & \underline{12.7} & 5.9 \\
 Qwen3-VL-8B-Instruct~\cite{qwen3vl} & 13.8 & 7.7 & 50.1 & 53.4 & 21.4 & 13.6 & 14.1 & 1.3 & 0.0 & 10.7 & 30.5 & 0.7 & 19.5 \\
 Qwen3-VL-32B-Instruct~\cite{qwen3vl} & 20.1 & 13.0 & 66.1 & 63.0 & 17.6 & 29.4 & 18.9 & 1.3 & 0.0 & 16.6 & 39.0 & 0.7 & 25.5 \\
 Qwen3-VL-235B-A22B-Instruct~\cite{qwen3vl} & 23.8 & 14.8 & \underline{68.9} & \underline{67.8} & \underline{32.9} & 28.3 & \underline{26.8} & 7.7 & 3.3 & 19.3 & \underline{45.0} & 5.5 & 30.5 \\
 GLM-4.5V~\cite{hong2025glm} & \textbf{59.1} & \textbf{45.4} & 66.4 & 61.0 & 21.2 & \underline{29.8} & 15.3 & 11.0 & 3.3 & \textbf{52.3} & 38.7 & 7.2 & \underline{34.7} \\
  InternVL-3.5-8B~\cite{wang2025internvl3} & 15.2 & 9.6 &25.9 & 42.4 & 6.1 & 9.8 & 3.1 & 0.0 & 0.0 & 12.4 & 17.5 & 0.0 &12.5 \\
 InternVL-3.5-38B~\cite{wang2025internvl3} & 28.0 & 15.9 &47.0 & 54.5 & 9.6 & 18.2 & 12.4 & 0.0 & 0.0 & 21.9 & 28.3 & 0.0 &20.6 \\

\midrule
\rowcolor{ourscolor}    \textbf{ VideoSSR-8B (Ours)} & \underline{29.0} & \underline{24.6} & \textbf{88.7} & \textbf{89.0} & \textbf{94.4} & \textbf{67.8} & \textbf{41.0} & \textbf{24.3} & \textbf{8.0} & \underline{26.8} & \textbf{76.2} & \textbf{16.2} & \textbf{51.9} \\
\bottomrule
\end{tabular}%
}
\end{table*}

\noindent\textbf{Anomaly Grounding.}
For the Anomaly Grounding task, we select five representative perturbation types from a larger pool of 14 (detailed in Appendix~\ref{subsec:example_anomaly_grounding}). The selected types include: (1) swapping the red and blue color channels, 
(2) rotation by 180 degrees, (3) zooming out, (4) horizontal mirroring, 
and (5) shuffling the intra-segment frame order.

We compute the Mean Intersection over Union (mIoU) between the predicted and ground-truth temporal intervals as the performance score.

\vspace{5pt}
\noindent\textbf{Object Counting.}
For the Object Counting task, we use three primitive shapes (circles, rectangles, and triangles) and configure two difficulty levels:
\begin{itemize}
    \item \textbf{Easy}: Objects are overlaid onto a maximum of three frames, with no more than three instances of any single shape type appearing in any given frame.
    \item \textbf{Hard}: The constraints are increased to a maximum of four frames and up to four instances per shape per frame.
\end{itemize}
A score of 1 is awarded for a specific shape type if the predicted count is exactly equal to the ground-truth count, and 0 otherwise. The final task score is the average of these binary scores across all shape types.

\vspace{5pt}
\noindent\textbf{Temporal Jigsaw.}
The Temporal Jigsaw task is configured with two difficulty settings based on the number of segments the video is partitioned into:
\begin{itemize}
    \item \textbf{Easy}: The video is partitioned into 6 segments.
    \item \textbf{Hard}: The video is partitioned into 8 segments.
\end{itemize}
 A score is awarded only if the predicted sequence of segments is identical to the ground-truth permutation.

As shown in Table~\ref{tab:viubench}, we evaluate a suite of powerful MLLMs on VIUBench. 
Our findings reveal that this benchmark poses a significant challenge even for the most advanced models. 
Notably, even a strong closed-source model like GPT-5~\cite{OpenAI2025-GPT5} only achieves a modest average score of \textbf{58.7}. 
The performance of open-source models is even more limited. For instance, Qwen3-VL-8B attains an average score of just 19.5.
These results underscore a critical insight that understanding and reasoning about intrinsic video properties, such as fine-grained details and temporal coherence, remains a substantial bottleneck for current MLLMs. 
This highlights the effectiveness of VIUBench in exposing the limitations of existing models and validates its role as a challenging benchmark for future research.

More importantly, our experiments with VIUBench reveal a key advantage of these pretext tasks: the difficulty of the generated questions can be easily scaled by adjusting simple parameters. 
For instance, in the Object Counting task, switching from the ``Easy" to the ``Hard" configuration caused the score of GPT-5 to drop sharply from 88.4 to 70.3.
A similar trend is observed in the Temporal Jigsaw task. 
By increasing the number of video segments from six to eight, the model's score plummeted from 39.0 to 27.0. 
Even for Video Jigsaw~\cite{wu2025visual}, a model specifically trained on jigsaw tasks, its performance decreases significantly from 20.3 to 5.0 under the same conditions.
To sum up, all these findings demonstrate that our method can dynamically generate tasks that challenge powerful MLLMs. The ability to parametrically control task difficulty ensures that VIUBench can remain a relevant and challenging benchmark for evaluating the continuous advancements of future models.

\begin{table*}[t]
\centering
\caption{\textbf{Performance comparison on General Video QA and Long Video QA tasks.}}
\label{tab:main_results1}
\resizebox{\textwidth}{!}{%
\begin{tabular}{lc cccc cccc}
\toprule
& & \multicolumn{4}{c}{\textbf{General Video QA}} & \multicolumn{4}{c}{\textbf{Long Video QA}} \\
\cmidrule(lr){3-6} \cmidrule(lr){7-10}

\textbf{Model} & \textbf{Frames} & \textbf{MVBench} & \textbf{TempCompass} & \textbf{AoTBench} & \textbf{VinoGround} & \textbf{Video-MME} & \textbf{LVBench} & \textbf{LongVideoBench} & \textbf{CGBench} \\
\midrule

\multicolumn{10}{c}{\cellcolor{separatorcolorclose}\textit{Closed Source Models}} \\
\midrule
GPT-4o~\cite{gpt4o} & - & 64.6 & 73.8 & 63.4 & 54 & 71.9 & 30.8 & 62.0 & 45.2 \\
Gemini-1.5-Pro~\cite{gemini1.5} & - & 60.5 & 67.1 & 58.3 & 35.8 & 75.0 & 33.1 & 58.6 & 37.2 \\
\midrule

\multicolumn{10}{c}{\cellcolor{separatorcolor}\textit{Open Source Models}} \\
\midrule
& 32 & 66.9 & 74.9 & 59.8 & 45.2 & 64.1 & 39.6 & 58.6 & 40.1 \\
   Qwen3-VL-8B-Instruct~\cite{qwen3vl}          & 48 & 67.6 & 74.8 & 60.2 & 45.0 & 66.0 & 41.5 & 60.0 & 41.7 \\
            & 64 & 67.8 & 74.8 & 60.7 & 45.0 & 66.4 & 43.0 & 61.3 & 42.6 \\
\midrule

\rowcolor{ourscolor}  & 32 & 68.6{(+1.7)} & 75.7{(+0.8)} & 60.5{(+0.7)} & 55.6{(+10.4)} & 65.2{(+1.1)} & 41.5{(+1.9)} & 59.6{(+1.0)} & 40.2{(+0.1)} \\
\rowcolor{ourscolor}                 \textbf{VideoSSR-8B (Ours)}         & 48 & 68.8{(+1.2)} & 75.8{(+1.0)} & 61.7{(+1.5)} & 55.6{(+10.6)} & 66.7{(+0.7)} & 42.9{(+1.4)} & 61.1{(+1.1)} & 42.5{(+0.8)} \\
\rowcolor{ourscolor}                          & 64 & 68.9{(+1.1)} & 75.7{(+0.9)} & 61.8{(+1.1)} & 55.6{(+10.6)} & 67.6{(+1.2)} & 44.0{(+1.0)} & 61.5{(+0.2)} & 43.4{(+0.8)} \\
\bottomrule
\end{tabular}
}
\end{table*}

\subsection{Video Self-Supervised Reinforcement Learning}
\label{sec:videossr}

Motivated by the insights from our proposed VIUBench, 
we introduce a novel framework that leverages \textbf{Video S}elf-\textbf{S}upervised \textbf{R}einforcement learning (\textbf{VideoSSR}) to enhance the generalization of MLLMs. 
To perform reinforcement learning, we first construct the \textbf{VideoSSR-30K} dataset. 
Specifically, this dataset consists of the aforementioned three  pretext tasks. 
The proportional distribution of VideoSSR-30K dataset is detailed in the right panel of Figure~\ref{fig:data}.

For training, we employ RLVR using GRPO~\cite{deepseekmath,deepseekr1}. We do not use recent variants of GRPO~\cite{huang2025spotlight,yu2025dapo,gspo}, as our primary focus is on the data itself. 

Our reward function is based solely on answer correctness. While these tasks are designed to be difficult, this very characteristic poses a problem for RLVR training: using a strict reward function often results in sparse rewards, leading to inefficient and unstable training.

To address this challenge, we design a specific smooth reward function for each task to provide a denser and more informative learning signal.

\vspace{5pt}
\noindent\textbf{Anomaly Grounding.}
For the temporal grounding of anomalies, the Mean Intersection over Union (mIoU) naturally serves as a smooth reward signal. It provides a score between 0 and 1 that reflects the degree of overlap between the predicted and ground-truth temporal segments. Let $T_{\text{pred}}$ and $T_{\text{gt}}$ be the predicted and ground-truth intervals, respectively. The reward $R_{\text{ground}}$ is simply as below:
\begin{equation}
    R_{\text{ground}} = \text{IoU}(T_{\text{pred}}, T_{\text{gt}}) = \frac{|T_{\text{pred}} \cap T_{\text{gt}}|}{|T_{\text{pred}} \cup T_{\text{gt}}|}
    \label{eq:reward_ground}
\end{equation}

\vspace{5pt}
\noindent\textbf{Object Counting.}
For the counting task, our reward function provides a dense signal based on the average relative error across all shape categories. For each category $k$, we first compute a score $R_{\text{count}, k}$ that is inversely proportional to the relative error. Let $y_k$ be the ground-truth count and $\hat{y}_k$ be the predicted count for category $k$. The score for a single category is:
\begin{equation}
    R_{\text{count}, k} = \max\left(0, 1 - \frac{|\hat{y}_k - y_k|}{y_k + \varepsilon}\right)
    \label{eq:reward_count_single_detailed}
\end{equation}
Here, the absolute error is normalized by the magnitude of the ground-truth value, and a small constant $\varepsilon$ (e.g., $10^{-9}$) ensures numerical stability. The final reward for the entire task, $R_{\text{count}}$, is the average of these scores over all $K$ shape categories:
\begin{equation}
    R_{\text{count}} = \frac{1}{K} \sum_{k=1}^{K} R_{\text{count}, k}
    \label{eq:reward_count_final}
\end{equation}

\vspace{5pt}
\noindent\textbf{Temporal Jigsaw.}
For the jigsaw puzzle, our reward function measures the structural correctness of the predicted sequence. We compute a penalty based on the cumulative displacement of elements from their correct positions. 
Let $P_{gt}$ be the ground-truth permutation and $\hat{P}$ be the predicted permutation. Let $\text{pos}(v, P)$ denote the position of an element $v$ in a sequence $P$. The total displacement error $E_{\text{jigsaw}}$ is defined as:
\begin{equation}
    E_{\text{jigsaw}} = \sum_{k=1}^{n} |\text{pos}(k, \hat{P}) - \text{pos}(k, P_{gt})|
\end{equation}
This error is then normalized by the maximum possible error, $E_{\max}$, which occurs for a reversed sequence. The final reward is given by:
\begin{equation}
    R_{\text{jigsaw}} = 1 - \frac{E_{\text{jigsaw}}}{E_{\max}}
    \label{eq:reward_jigsaw}
\end{equation}

%% file: sec/4_experiment.tex
\begin{table*}[t]
\centering
\caption{\textbf{Performance comparison on Temporal Grounding and Complex Reasoning tasks.}}
\label{tab:main_results2}
\resizebox{\textwidth}{!}{%
\begin{tabular}{lc cccc cccc}
\toprule
& & \multicolumn{4}{c}{\textbf{Temporal Grounding}} & \multicolumn{4}{c}{\textbf{Complex Reasoning}} \\
\cmidrule(lr){3-6} \cmidrule(lr){7-10}

\textbf{Model} & \textbf{Frames}  & \textbf{QVHighlights} & \textbf{ActivityNet}& \textbf{CharadesSTA} & \textbf{TACoS} & \textbf{VideoMMMU} & \textbf{Video-TT} & \textbf{VCRBench} & \textbf{CVBench} \\
\midrule

\multicolumn{10}{c}{\cellcolor{separatorcolorclose}\textit{Closed Source Models}} \\
\midrule
GPT-4o~\cite{gpt4o} &  - &- &  -& 35.7 &- & 61.2 & 45.2 & 29.0 & 69.2 \\
Gemini-1.5-Pro~\cite{gemini1.5} & - &- & - & - & - & 53.9 & 38.2 & 48.2 & - \\
\midrule

\multicolumn{10}{c}{\cellcolor{separatorcolor}\textit{Open Source Models}} \\
\midrule
& 32 &  43.7 & 36.5 &50.3 & 22.4 & 58.2 & 41.8 & 7.4 & 61.8 \\
   Qwen3-VL-8B-Instruct~\cite{qwen3vl}          & 48 &  46.4 &38.4 &50.0 & 25.9 & 58.5 & 43.0 & 7.4 & 61.5 \\
            & 64 & 48.6 & 39.8 &49.2& 28.1 &58.8 & 44.0 & 8.8 & 61.6 \\
\midrule

\rowcolor{ourscolor}  & 32 &  59.6{(+15.9)} & 42.1{(+5.6)} &52.1{(+1.8)} & 23.1{(+0.7)} & 59.9{(+1.7)} & 44.2{(+2.4)} & 10.7{(+3.3)} & 63.5{(+1.7)} \\
\rowcolor{ourscolor}                 \textbf{VideoSSR-8B (Ours)}         & 48  & 61.1{(+14.7)} & 43.0{(+4.6)} & 51.1{(+1.1)}& 27.7{(+1.8)} & 60.0{(+1.5)} & 44.9{(+1.9)} & 15.3{(+7.9)} & 63.8{(+2.3)} \\
\rowcolor{ourscolor}                          & 64 &  62.6{(+14.0)} &43.7{(+3.9)} &49.9{(+0.7)} & 30.6{(+2.5)} & 60.9{(+2.1)} & 45.8{(+1.8)} & 17.8{(+9.0)} & 63.3{(+1.7)} \\
\bottomrule
\end{tabular}
}
\end{table*}

\section{Experiments}
\paragraph{Implementation Details.}

In this paper, our VideoSSR model is built upon the Qwen3-VL-8B-Instruct~\cite{qwen3vl}. 
We perform RLVR on our newly constructed VideoSSR-30K dataset for one epoch.
Key hyperparameters for training include a learning rate of $1 \times 10^{-6}$, a global batch size of 64, and a rollout number ($N$) of 8 for generation, a KL divergence penalty with a coefficient of $1 \times 10^{-3}$. MAX\_FRAMES is configured to 48, and MAX\_PIXELS is set to $256 \times 256$ for efficient training. 
The entire training process is conducted on 8 H200 GPUs and takes approximately 16 hours.
{To ensure a fair and reproducible comparison, both Qwen3-VL and VideoSSR are evaluated under identical conditions: FPS is set to 2, with MAX\_FRAMES configured to \{32, 48, 64\}. MAX\_PIXELS is set to $512 \times 512$. Greedy decoding is used to ensure reproducibility. Chain of thought~\cite{wei2022chain} is not utilized to mitigate hallucination~\cite{luo2025thinking} and ensure correct output formatting, therefore enhancing performance.}

\paragraph{Benchmarks and Baselines.}
To comprehensively evaluate the generalization capability of VideoSSR, we conduct experiments on 16 distinct benchmarks spanning four major video task categories:

\begin{itemize}
    \item \textbf{General Video QA:} MVBench~\cite{li2024mvbench}, TempCompass~\cite{liu2024tempcompass}, AoTBench~\cite{aot}, and VinoGround~\cite{zhang2024vinoground}.
    \item \textbf{Long Video QA:} Video-MME~\cite{fu2025video}, LVBench~\cite{wang2024lvbench}, LongVideoBench~\cite{wu2024longvideobenchbenchmark}, and CGBench~\cite{chen2024cgbench}.
    \item \textbf{Temporal Grounding:}  QVHighlights~\cite{lei2021detecting}, ActivityNet~\cite{caba2015activitynet}, CharadesSTA~\cite{gao2017tall}, and TACoS~\cite{regneri2013grounding}.
    \item \textbf{Complex Reasoning:} VideoMMMU~\cite{hu2025video}, Video-TT~\cite{zhang2025towards}, VCRBench~\cite{sarkar2025vcrbench}, and CVBench~\cite{zhu2025cvbench}.
\end{itemize}
For the Temporal Grounding tasks, we report the Mean Intersection over Union (mIoU) as the primary evaluation metric. 
Further details regarding each benchmark and a full breakdown of the results can be found in Appendix~\ref{subsubsec:more_results}.

For our primary baseline, we select Qwen3-VL-8B-Instruct, as it represents the state-of-the-art among open-source models. 
To further contextualize the performance of our method, we also provide a comparative analysis against two formidable proprietary models: GPT-4o~\cite{gpt4o} and Gemini-1.5-Pro~\cite{gemini1.5}.

\subsection{Main Results}
\noindent\textbf{General Video QA}
As shown in the left half of Table~\ref{tab:main_results1}, VideoSSR achieves substantial improvements on temporally related benchmarks such as VinoGround~\cite{zhang2024vinoground}, even surpassing closed source models. It also obtains improvements on more general benchmarks, for instance on 
MVBench~\cite{li2024mvbench}, achieving a score of 68.9 and similarly outperforming the closed source models.

\vspace{5pt}
\noindent\textbf{Long Video QA}
As shown in the right half of Table~\ref{tab:main_results1}, VideoSSR also achieves consistent improvements on four mainstream benchmarks. Because we primarily conduct training and evaluation with a low number of frames, a gap remains compared to closed-source models on such long video understanding tasks, which is a direction for future research.

\vspace{5pt}
\noindent\textbf{Temporal Grounding}
As shown in the left side of Table~\ref{tab:main_results2}, benefiting from the Anomaly Grounding task, VideoSSR achieves remarkable zero-shot improvements on multiple mainstream temporal grounding benchmarks, especially on QVHighlights~\cite{lei2021detecting} and ActivityNet~\cite{caba2015activitynet}, with gains of +15.9 and +5.6, respectively.

\vspace{5pt}
\noindent\textbf{Complex Reasoning}
As shown in the right side of Table~\ref{tab:main_results2}, VideoSSR achieves a large improvement of +9.0 on VCRBench~\cite{sarkar2025vcrbench}, a benchmark that is highly correlated with our Temporal Jigsaw task. It also obtains consistent improvements on other video reasoning benchmarks.

\vspace{5pt}
In summary, we validate the generalization capability of VideoSSR on the 16 aforementioned benchmarks. Notably, VideoSSR achieves consistent performance improvements across four major video tasks under three different frame settings. Under the 48 frame setting, VideoSSR obtains an average improvement of 5.1\% across all 17 benchmarks (including VIUBench), comprehensively demonstrating the effectiveness of VideoSSR.


\subsection{Ablation Study}
\label{sec:ablation_study}

    
    
\paragraph{Analysis on three pretext tasks.} First, we individually validate the effectiveness of the three pretext tasks, as shown in Table~\ref{tab:ablation_tasks}. Benefiting from its design, the Anomaly Grounding task leads to a significant performance increase on CharadesSTA. Similarly, the Temporal Jigsaw task brings a substantial boost to VCRBench. Notably, all three tasks individually improve performance on Video-MME, confirming their contribution to enhancing general video understanding capabilities.

\begin{table}[h]
\centering
\caption{\textbf{Ablation study of the three pretext tasks and their corresponding smooth reward functions.} G, C, and J represent Anomaly Grounding, Object Counting, and Temporal Jigsaw, respectively. $\checkmark$ indicates the component is used for training. R@0.5 denotes recall at an IoU threshold of 0.5. Step denotes step accuracy for VCRBench. The best and second best results are shown in bold and underlined.}

\label{tab:ablation_tasks}
\resizebox{\columnwidth}{!}{%
\begin{tabular}{ccc|c|cc|cc|cc} 
\hline
\multicolumn{4}{c|}{\textbf{Training Config}} & \multicolumn{2}{c|}{\textbf{Understanding}} & \multicolumn{2}{c|}{\textbf{Grounding}} & \multicolumn{2}{c}{\textbf{Reasoning}} \\ \hline
\multicolumn{3}{c|}{\textbf{Pretext Tasks}} & \textbf{Reward} & \multicolumn{2}{c|}{\textbf{Video-MME}} & \multicolumn{2}{c|}{\textbf{CharadesSTA}} & \multicolumn{2}{c}{\textbf{VCRBench}} \\ \hline
 G & C & J & Smooth & All & Long & mIoU & R@0.5 & Acc & Step  \\ \hline
\rowcolor[HTML]{E6E6FA} \multicolumn{10}{c}{\textit{Baseline Model}} \\ \hline
\ding{55} & \ding{55} & \ding{55} & -- & 64.1 & 54.3 & 50.3 & 58.4 & 7.4 &25.9 \\ \hline
\rowcolor[HTML]{E6E6FA} \multicolumn{10}{c}{\textit{Models on Subtasks}} \\ \hline
$\checkmark$ & \ding{55} & \ding{55} & \ding{55} & 64.7 & 54.8 & 47.5 & 52.2 & 5.8 & 24.9 \\
$\checkmark$ & \ding{55} & \ding{55} & $\checkmark$ & 64.8 & 55.9 & \textbf{53.8} & \textbf{63.8} & 4.1 & 22.8 \\\hline
\ding{55} & $\checkmark$ & \ding{55} & \ding{55} & 64.7 & 54.7 & 51.5 & 59.9 & 6.3 & 25.4 \\
\ding{55} & $\checkmark$ & \ding{55} & $\checkmark$ & \underline{64.9} & \underline{56.2} & 51.4 & 60.1 & 5.5 & 24.8 \\ \hline
\ding{55} & \ding{55} & $\checkmark$ & \ding{55} & 64.3 & 55.0 & 51.3 & 59.1 & \underline{13.4} & \underline{32.7} \\
\ding{55} & \ding{55} & $\checkmark$ & $\checkmark$ & 64.8 & 55.8 & 51.0 & 59.0 & \textbf{15.9} & \textbf{35.5} \\ \hline
\rowcolor[HTML]{E6E6FA} \multicolumn{10}{c}{\textit{Models on All Tasks}} \\ \hline
$\checkmark$ & $\checkmark$ & $\checkmark$ & \ding{55} & 64.8 & 55.4 & 51.3 & 59.3 & 10.7 & 30.4 \\ 
$\checkmark$ & $\checkmark$ & $\checkmark$ & $\checkmark$ & \textbf{65.2} & \textbf{57.1} & \underline{52.1} & \underline{60.6} & 10.7 & 32.3 \\ \hline
\end{tabular}%
}
\end{table}

Moreover, Table~\ref{tab:ablation_tasks} also shows the impact of the smooth reward function on the results. 
We observe that the model trained with a strict matching reward function performs closer to the baseline. This is because a strict reward function often leads to sparse reward signals, which are more likely to result in a zero advantage in GRPO. Consequently, the training becomes inefficient, leading to smaller update magnitudes.
Furthermore, this 
approach introduces training instability. For instance, training the anomaly grounding 
task with a strict reward function even degrades performance on CharadesSTA.

\begin{figure}[h!]
    \centering
    \includegraphics[width=\columnwidth]{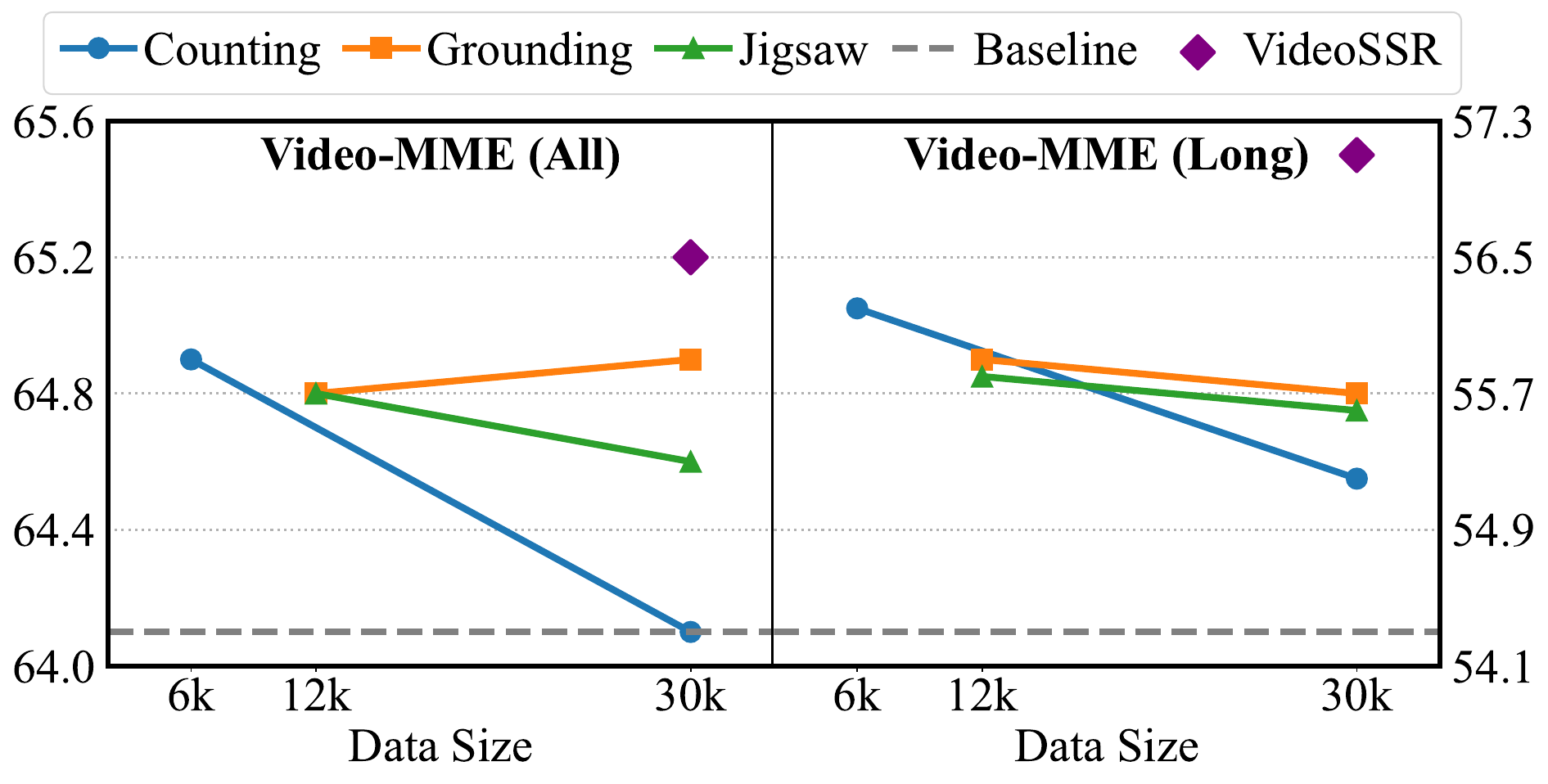}
    
    \caption{\textbf{Comparison of single task and mixed task training at the 30k data scale.} 
The results demonstrates that task diversity is more effective for improving performance than simply scaling up the data for a single pretext task.}
\label{fig:ablation_task_diversity}
\end{figure}

To further investigate the benefits of task diversity, we conducted a comparative analysis between single task training and our mixed task VideoSSR-30K framework, controlling for the data scale at 30k samples for both settings. As illustrated in Figure~\ref{fig:ablation_task_diversity}, we observe that simply scaling up the data for a single task yields diminishing returns and even degrades performance. This finding suggests that designing a diverse and rich set of pretext tasks, rather than focusing on a single one, is a more promising direction for enhancing model capabilities.

We also compare our method against training with LongVideoReason~\cite{chen2025longvila-r1} or ReWatch~\cite{rewatch-r1}. We utilize only the multiple-choice subsets from each dataset. The specific training procedures are:

\begin{itemize}
    \item For \textbf{LongVideoReason}, the model is trained for 500 steps with a batch size of 64.
    \item For \textbf{ReWatch}, we use a composite subset of questions from its Video-R1~\cite{feng2025videor1} and VideoEspresso~\cite{han2025videoespresso} portions and train the model for one full epoch.
\end{itemize}
The results are presented in Table~\ref{tab:ablation_data}. Notably, the model trained with VideoSSR-30K surpasses the performance of models trained on annotated datasets of a comparable scale. Furthermore, we observe a critical limitation: fine-tuning the powerful Qwen3-VL on LongVideoReason, a dataset annotated by a less capable MLLM, can even lead to performance degradation, which further demonstrate the importance of our self-supervised paradigm.

\begin{table}[h]
\centering
\caption{\textbf{Ablation study on different training datasets. ``None'' indicates the baseline Qwen-VL3.}}
\label{tab:ablation_data}
\resizebox{\columnwidth}{!}{%
\begin{tabular}{l|c|cc|cc|cc} 
\hline
\multicolumn{2}{c|}{\textbf{Training Config}} & \multicolumn{2}{c|}{\textbf{Video-MME}} & \multicolumn{2}{c|}{\textbf{CharadesSTA}} & \multicolumn{2}{c}{\textbf{VCRBench}} \\
\hline 
\textbf{Fine-tuning Data} & \textbf{Size} & All & Long & mIoU & R@0.5 & Acc & Step  \\ \hline
\rowcolor[HTML]{E6E6FA} \multicolumn{8}{c}{\textit{Baseline Model}} \\ \hline
None & -- & 64.1 & 54.3 & 50.3 & 58.4 & 7.4 & 25.9 \\
 \hline
 \rowcolor[HTML]{E6E6FA} \multicolumn{8}{c}{\textit{Fine-tuned Models}} \\ \hline
 LongVideoReason~\cite{chen2025longvila-r1} & 32k & 63.6 & 53.3 & 51.7 & 59.4 & 7.1 & 26.1 \\
 ReWatch~\cite{rewatch-r1} & 27k & 64.7 & 56.7 & 51.6 & 59.2 & 2.7 & 22.2 \\ \hline
\textbf{VideoSSR-30K} & \textbf{30K} & \textbf{65.2} & \textbf{57.1} & \textbf{52.1} & \textbf{60.6} & \textbf{10.7} & \textbf{32.3} \\ \hline
\end{tabular}%
}
\end{table}

Finally, we explored the optimal selection of subtasks for the Anomaly Grounding task. 
We investigate 14 distinct perturbation types and report their corresponding accuracies 
on Video-MME, as illustrated in Figure~\ref{fig:ab_grounding}. 
Further 
details are provided in Appendix~\ref{subsec:example_anomaly_grounding}. Based on these results, 
we select four perturbations that offer substantial improvements and create a 
uniform mixture to construct our final training set.
Furthermore, we found that perturbations targeting temporal properties, 
such as simulating a fast forward effect by sampling denser frame sequences, does not 
appear to yield benefits and even introduced negative side effects. 
This may be because 
the base model, Qwen3-VL, relies on textual timestamps for its temporal awareness. 
Deliberately creating visual anomalies in this domain might confuse the model rather than enhance its learning.
\begin{figure}[h!]
    \centering
    \includegraphics[width=\columnwidth]{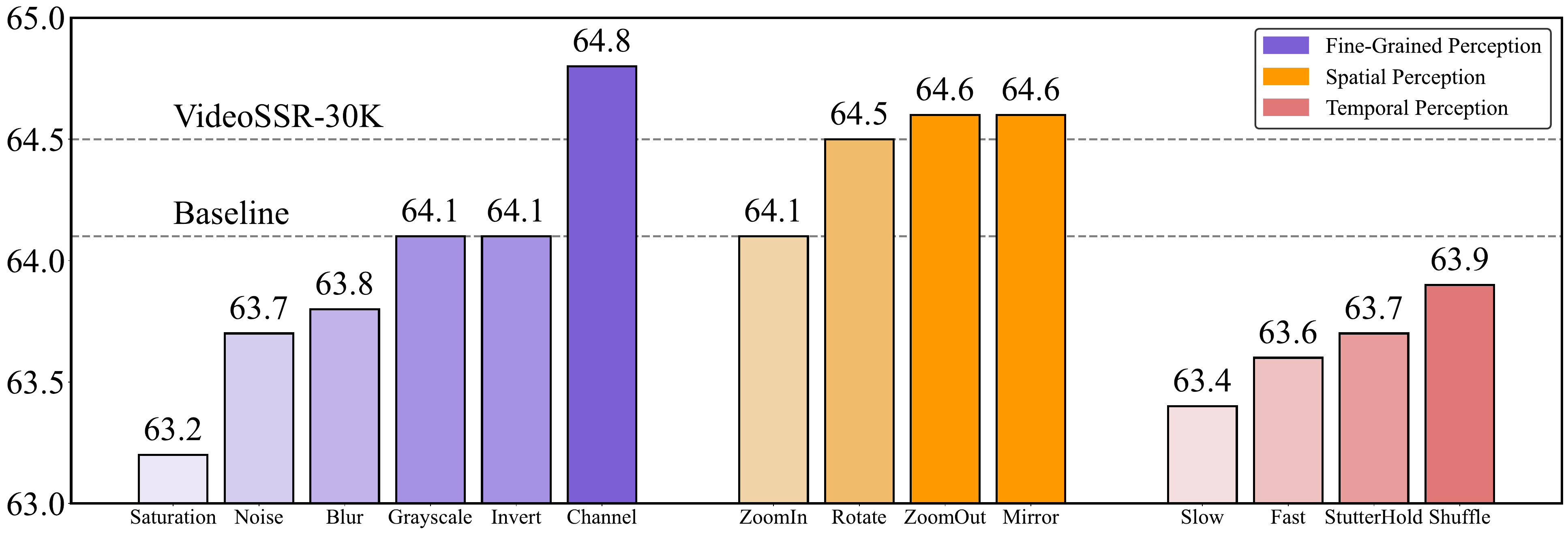}
    
    \caption{\textbf{Ablation study of the 14 perturbation subtypes for Anomaly Grounding.} Accuracy is reported on Video-MME.}
    
    \label{fig:ab_grounding} 
\end{figure}

%% file: sec/5_conclusion.tex
\section{Conclusion}
\label{sec:conclusion}

In this paper, we introduce VideoSSR, a novel self-supervised reinforcement learning framework designed to address the critical limitations of existing video datasets for training MLLMs. 
By designing the three pretext tasks of Anomaly Grounding, Object Counting, and Temporal Jigsaw, we construct 
the challenging VIUBench and the VideoSSR-30K dataset without reliance on manual or MLLM annotations.
Our extensive experiments demonstrate that VideoSSR leads to consistent and significant performance 
gains across 17 diverse benchmarks, including four main video tasks, achieving an average improvement of over 5\%.
Our work highlights self-supervision as a powerful method for generating scalable, low-cost, and high-quality training data. 
Crucially, the parametric control over task difficulty ensures 
the long-term relevance of our framework for benchmarking increasingly capable MLLMs. 
Our approach moves beyond the limitations of static, annotated datasets, enabling the development of models that learn directly from the intrinsic structure of video.

%% file: sec/X_suppl.tex
\clearpage
\clearpage
\appendix 

\begin{strip} 
    \centering 
    {\Huge\bfseries Supplementary Materials\par}
    \vspace{0.5cm}
\end{strip}

\section{Implementation Details}
\label{sec:implementation_details}

\subsection{Training Details}
\label{subsec:training_details}
We utilize Llava-Video~\cite{zhang2024video} as the primary video source for constructing both VideoSSR-30K dataset and VIUBench. 
During training, we did not employ chain of thought~\cite{wei2022chain} for VideoSSR or any models in the ablation studies. This decision aligns with our focus on enhancing fundamental perceptual abilities, namely, Fine-Grained, Spatial, and Temporal Perception, rather than complex reasoning. This approach also yields greater training efficiency and reduces the potential for model hallucination~\cite{luo2025thinking}.

\subsection{Evaluation Details for VideoSSR}
\label{subsec:evaluation_details}
\subsubsection{Prompts}

\begin{figure}[h!]
    \centering
    \includegraphics[width=1\columnwidth]{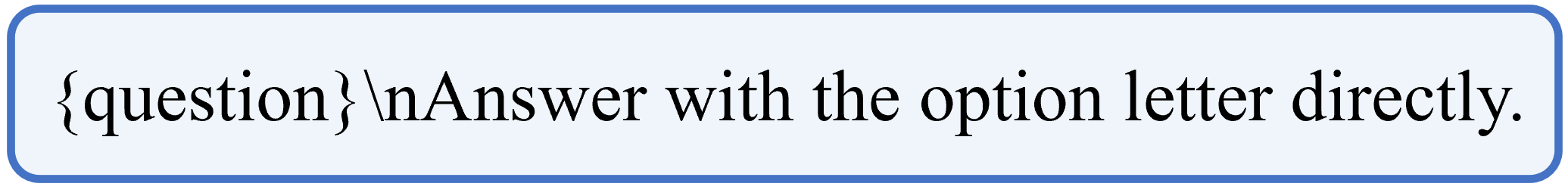}
    \caption{\textbf{Prompt template for Video QA tasks}}
    \label{fig:prompt_qa}
\end{figure}

For Video QA tasks, we prompt the model to generate a direct answer. The specific prompt template utilized for these tasks is illustrated in Figure~\ref{fig:prompt_qa}.

\begin{figure}[h!]
    \centering
    \includegraphics[width=1\columnwidth]{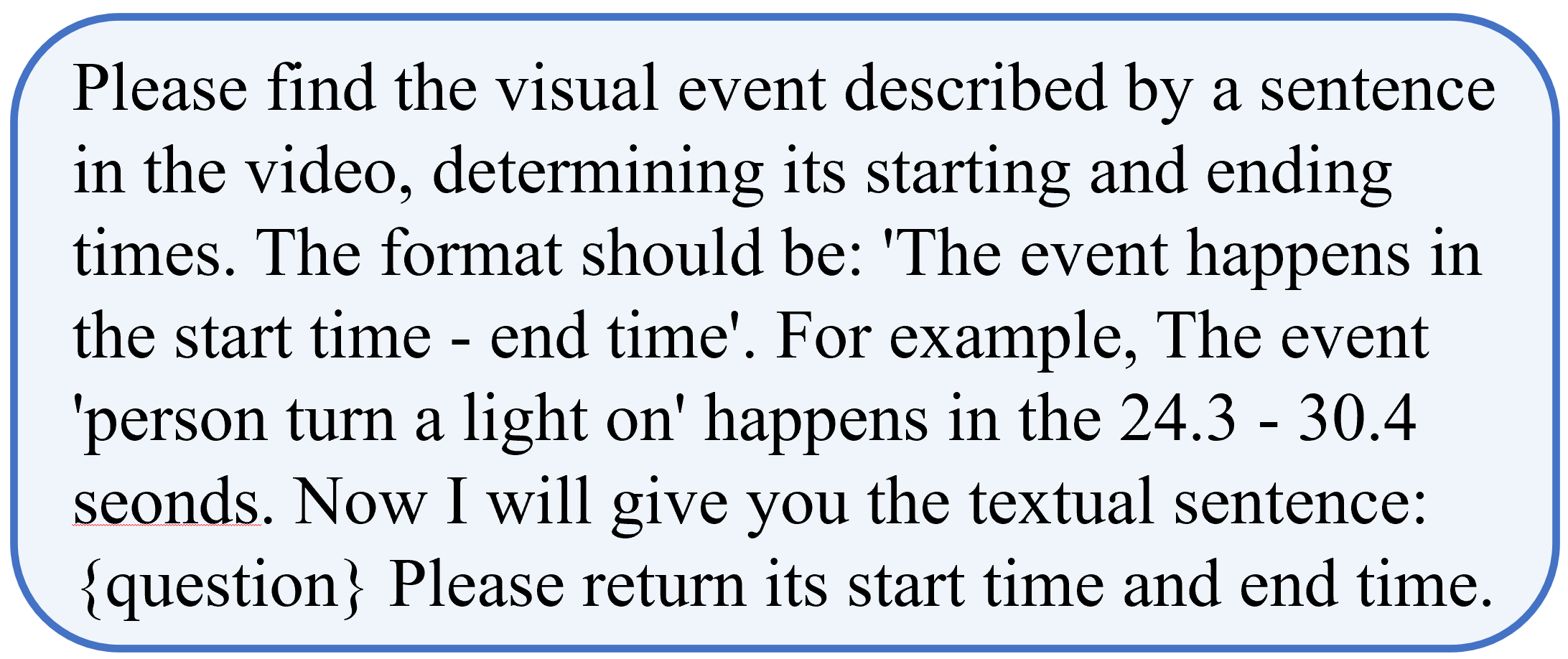}
    \caption{\textbf{Prompt template for Temporal Grounding tasks}}
    \label{fig:prompt_grounding}
\end{figure}

For Temporal Grounding tasks, our prompt format is based on the one utilized in the \texttt{lmms\_eval} library~\cite{zhang2024lmmsevalrealitycheckevaluation}, as depicted in Figure~\ref{fig:prompt_grounding}. While we observed that CharadesSTA seems to be particularly sensitive to prompt phrasing, we nonetheless applied this unified prompt across all benchmarks to ensure a fair and consistent evaluation.

For other specialized benchmarks, such as VCRBench, we adhere to the official prompts.

\subsubsection{Benchmarks}
We adhered to specific evaluation protocols for several benchmarks to ensure fair and accurate assessment.

\begin{itemize}
    \item \textbf{VinoGround:} We report the text score, which offers greater discriminative power between models.
    
    \item \textbf{Video-MME \& LongVideoBench:} For both benchmarks, evaluations are conducted without the use of subtitles. For LongVideoBench, we specifically test on its validation set.
    
    \item \textbf{CGBench:} Our evaluation is performed on its 3k subset.
    
    \item \textbf{Temporal Grounding:} For benehmarks in this category, the model is required to predict a single most likely temporal interval. Results for QVHighlights and ActivityNet are reported on their validation sets.
    
    \item \textbf{VideoMMMU \& Video-TT:} We report results on the multiple choice subset to facilitate answer extraction and comparison.
    
    \item \textbf{CVBench:} Our evaluation uses configurations of 32, 48, and 64 frames for each video, resulting in a significantly larger total number of frames processed per query.

\end{itemize}

\subsubsection{Detailed Results}
\label{subsubsec:more_results}
For Temporal Grounding tasks, we provide a more detailed breakdown of the results, as detailed in Table~\ref{tab:more_results} and Table~\ref{tab:more_results2}.
\renewcommand\theadfont{\small\bfseries} 

\begin{table*}[t]
\centering
\caption{\textbf{More results on QVHighlights and ActivityNet.}}
\label{tab:more_results}
\resizebox{\textwidth}{!}{%
\begin{tabular}{lc cccc cccc}
\toprule
& & \multicolumn{4}{c}{\textbf{QVHighlights}} & \multicolumn{4}{c}{\textbf{ActivityNet}} \\
\cmidrule(lr){3-6} \cmidrule(lr){7-10}

\textbf{Model} & \textbf{Frames}  & \textbf{mIoU} & \textbf{R@0.3}& \textbf{R@0.5} & \textbf{R@0.7}&\textbf{mIoU} & \textbf{R@0.3}& \textbf{R@0.5} & \textbf{R@0.7} \\
\midrule

& 32 & 43.7 & 62.3 & 42.5 & 24.2 & 36.5 & 52.3 & 34.5 & 18.3 \\
Qwen3-VL-8B-Instruct~\cite{qwen3vl} & 48 & 46.4 & 64.4 & 46.5 & 30.3 & 38.4 & 54.3 & 36.4 & 21.0 \\
& 64 & 48.6 & 64.5 & 48.6 & 33.9 & 39.8 & 55.6 & 38.6 & 23.0 \\
\midrule

\rowcolor{ourscolor} & 32 & 59.6{(+15.9)} & 83.3{(+21.0)} & 66.0{(+23.5)} & 43.4{(+19.2)} & 42.1{(+5.6)} & 63.0{(+10.7)} & 41.4{(+6.9)} & 21.5{(+3.2)} \\
\rowcolor{ourscolor} \textbf{VideoSSR-8B (Ours)} & 48 & 61.1{(+14.7)} & 83.5{(+19.1)} & 66.9{(+20.4)} & 48.3{(+18.0)} & 43.0{(+4.6)} & 63.2{(+8.9)} & 42.3{(+5.9)} & 22.7{(+1.7)} \\
\rowcolor{ourscolor} & 64 & 62.6{(+14.0)} & 83.7{(+19.2)} & 68.0{(+19.4)} & 49.7{(+15.8)} & 43.7{(+3.9)} & 63.3{(+7.7)} & 42.7{(+4.1)} & 24.2{(+1.2)} \\
\bottomrule
\end{tabular}%
}
\end{table*}

\begin{table*}[t]
\centering
\caption{\textbf{More results on CharadesSTA and Tacos.}}
\label{tab:more_results2}
\resizebox{\textwidth}{!}{%
\begin{tabular}{lc cccc cccc}
\toprule
& & \multicolumn{4}{c}{\textbf{CharadesSTA}} & \multicolumn{4}{c}{\textbf{Tacos}} \\
\cmidrule(lr){3-6} \cmidrule(lr){7-10}

\textbf{Model} & \textbf{Frames}  & \textbf{mIoU} & \textbf{R@0.3}& \textbf{R@0.5} & \textbf{R@0.7}&\textbf{mIoU} & \textbf{R@0.3}& \textbf{R@0.5} & \textbf{R@0.7} \\
\midrule

& 32 & 50.3 & 76.5 & 58.1 & 27.9 & 22.4 & 34.7 & 19.2 & 7.1 \\
Qwen3-VL-8B-Instruct~\cite{qwen3vl} & 48 & 50.0 & 76.6 & 56.1 & 26.9 & 25.9 & 39.0 & 24.0 & 10.7 \\
& 64 & 49.2 & 77.1 & 54.2 & 25.5 & 28.1 & 42.0 & 26.6 & 12.3 \\
\midrule

\rowcolor{ourscolor} & 32 & 52.1{(+1.8)} & 78.2{(+1.7)} & 60.6{(+2.5)} & 30.8{(+2.9)} & 23.1{(+0.7)} & 34.1{(-0.6)} & 19.8{(+0.6)} & 7.4{(+0.3)} \\
\rowcolor{ourscolor} \textbf{VideoSSR-8B (Ours)} & 48 & 51.1{(+1.1)} & 79.0{(+2.4)} & 59.9{(+3.8)} & 27.5{(+0.6)} & 27.7{(+1.8)} & 40.0{(+1.0)} & 24.7{(+0.7)} & 12.3{(+1.6)} \\
\rowcolor{ourscolor} & 64 & 49.9{(+0.7)} & 78.7{(+1.6)} & 57.6{(+3.4)} & 24.2{(-1.3)} & 30.6{(+2.5)} & 43.8{(+1.8)} & 28.1{(+1.5)} & 14.4{(+2.1)} \\
\bottomrule
\end{tabular}%
}
\end{table*}

\subsection{Evaluation Details for VIUBench}
\label{subsec:viubench_evaluation_details}
All evaluations on VIUBench utilized a fixed input of 48 frames with a maximum resolution of $512 \times 512$ pixels. 

\section{Details of Pretext Tasks}
\label{sec:pretext_task_examples}

\subsection{Anomaly Grounding}
\label{subsec:example_anomaly_grounding}
\begin{figure}[h!]
    \centering
    \includegraphics[width=\columnwidth]{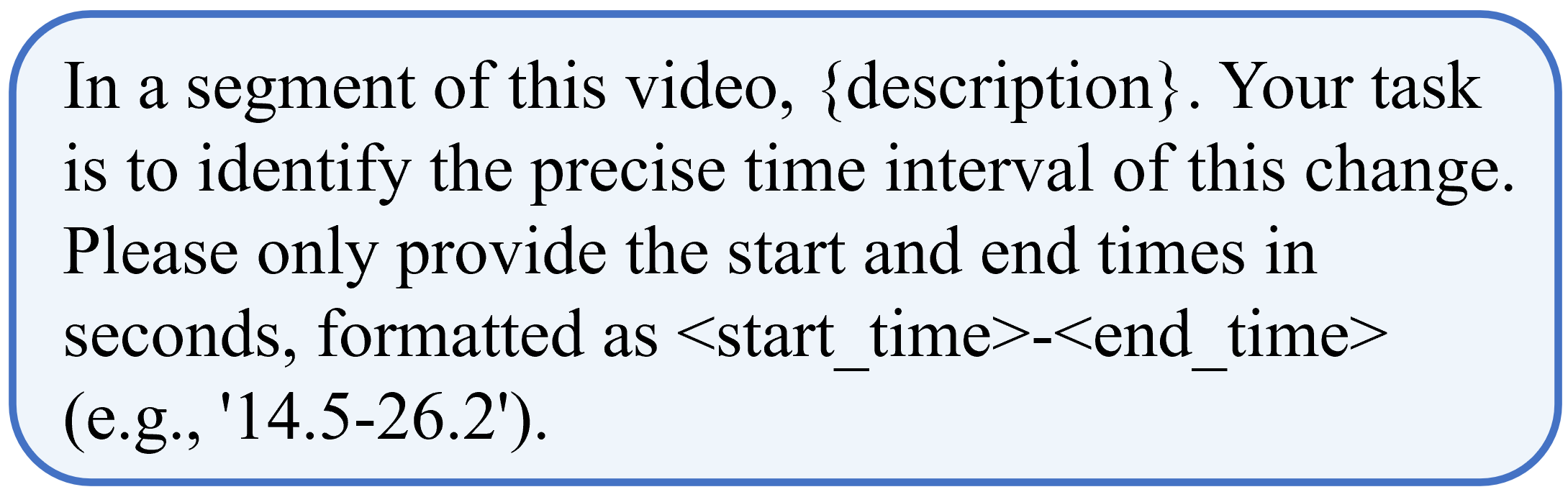} 
    \caption{\textbf{Prompt template for Anomaly Grounding.}}
    \label{fig:task_grounding_prompt}
\end{figure}

Figure~\ref{fig:task_grounding_prompt} illustrates the prompt template used for the Anomaly Grounding task. Table~\ref{tab:perturbation_details} provides the comprehensive list and definitions for all 14 perturbation subtypes designed for this task. The text in the ``Description" column of the table is what replaces the \texttt{\{description\}} placeholder in the prompt for each respective subtype.

\begin{table*}[t]
\centering
\caption{\textbf{Definitions of the 14 Perturbation Subtypes for Anomaly Grounding.} For temporal perception tasks, an additional detailed note (marked with *) was provided to guide the model.}
\label{tab:perturbation_details}
\begin{tabularx}{\textwidth}{@{}llX@{}}
\toprule
\textbf{Category} & \textbf{Perturbation Type} & \textbf{Description} \\
\midrule
\multicolumn{3}{l}{\textit{\textbf{Fine-Grained Perception}}} \\
\cmidrule(lr){1-3}
& Saturation & the colors in the video become oversaturated and unnaturally vibrant. \\
& Noise & Gaussian noise is added to the video. \\
& Blur & the video becomes blurry or out of focus. \\
& Grayscale & the video becomes black and white. \\
& Invert & the colors in the video are inverted. \\
& Channel Swap & the red and blue color channels in the video are swapped. \\
\midrule
\multicolumn{3}{l}{\textit{\textbf{Spatial Perception}}} \\
\cmidrule(lr){1-3}
& Zoom In & the video is zoomed in. \\
& Rotate & the video is rotated 180 degrees. \\
& Zoom Out & the video is zoomed out. \\
& Mirror & The video is mirrored horizontally. \\
\midrule
\multicolumn{3}{l}{\textit{\textbf{Temporal Perception}}} \\
\cmidrule(lr){1-3}
& Slow & the video slows down, this means the action unfolds at an unusually slow pace, making movements appear prolonged.* \\
& Fast & the video speeds up, this means the segment plays at a high speed, compressing the action and making movements appear jerky or rushed.* \\
& StutterHold & the video appears to freeze and stutter on a few frames, this means instead of playing smoothly, the video repeatedly freezes on a single frame before jumping to the next. \\
& Shuffle & the frames are shuffled, this means the order of events is scrambled, making the action appear illogical and chaotic. \\
\bottomrule
\multicolumn{3}{p{0.95\textwidth}}{\footnotesize \textbf{*Special Note for Slow/Fast perturbations:} To ensure a fair challenge, even if the video's actual speed changes (e.g., slow motion or fast forward), the timestamps for each frame have been intentionally kept evenly spaced. This creates the illusion of a constant playback speed. Therefore, you should not rely on the timestamps when judging the speed. Instead, your judgment must be based solely on the visual content. You should analyze the motion within the video itself by observing how much or how little the scene changes between consecutive frames to determine the true playback speed.} \\
\end{tabularx}%
\end{table*}

Notably, for perturbations targeting \textbf{Temporal Perception} (specifically \textit{Slow} and \textit{Fast}), we provided an expanded and highly detailed description within the prompt. This special note, as detailed at the bottom of Table~\ref{tab:perturbation_details}, explicitly instructed the model to disregard the evenly spaced frame timestamps and instead rely solely on visual motion cues. 

Despite this explicit guidance, the model's performance on these tasks remained notably poor, as shown in Figure~\ref{fig:ab_grounding}. We hypothesize that this is because the base model, Qwen3-VL, has a strong inherent bias towards relying on textual timestamp information when it is available. Forcing the model to overcome this bias and learn true visual motion perception appears to be a significant challenge, even with detailed and explicit prompting.

\begin{figure*}[t]
    \centering
    \includegraphics[width=\textwidth]{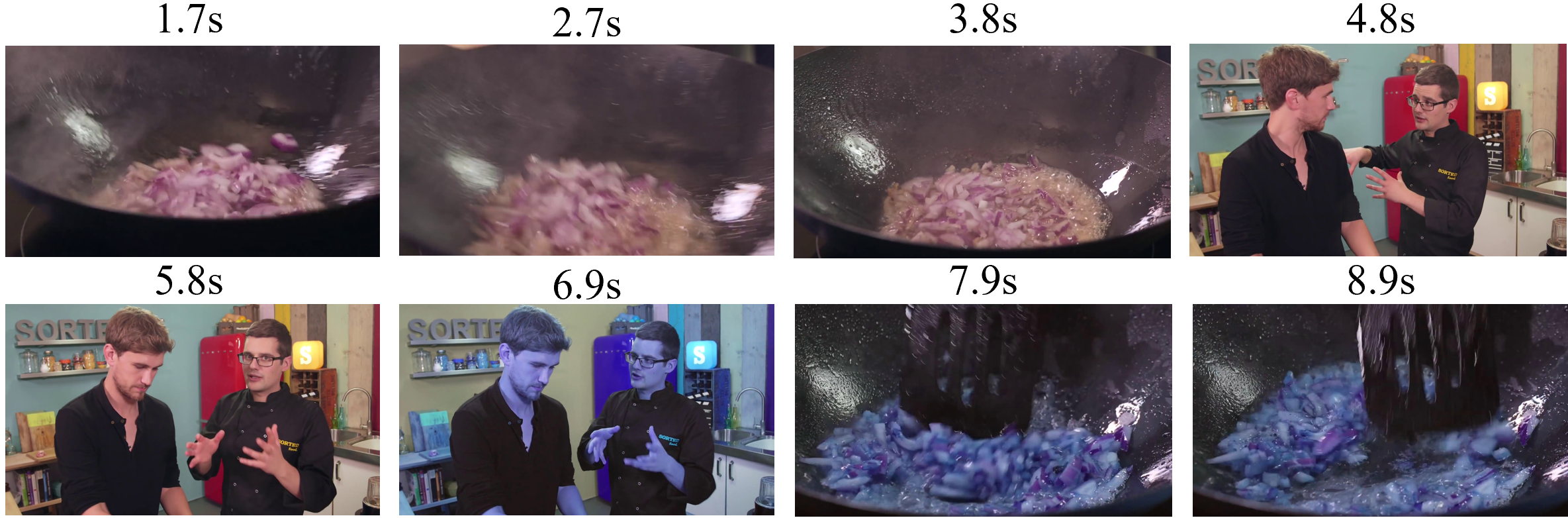}
    \caption{\textbf{An example of Channel Swap.} The ground truth is 6.9s--9.2s.}
    \label{fig:demo_channel}
\end{figure*}

\begin{figure*}[t]
    \centering
    \includegraphics[width=\textwidth]{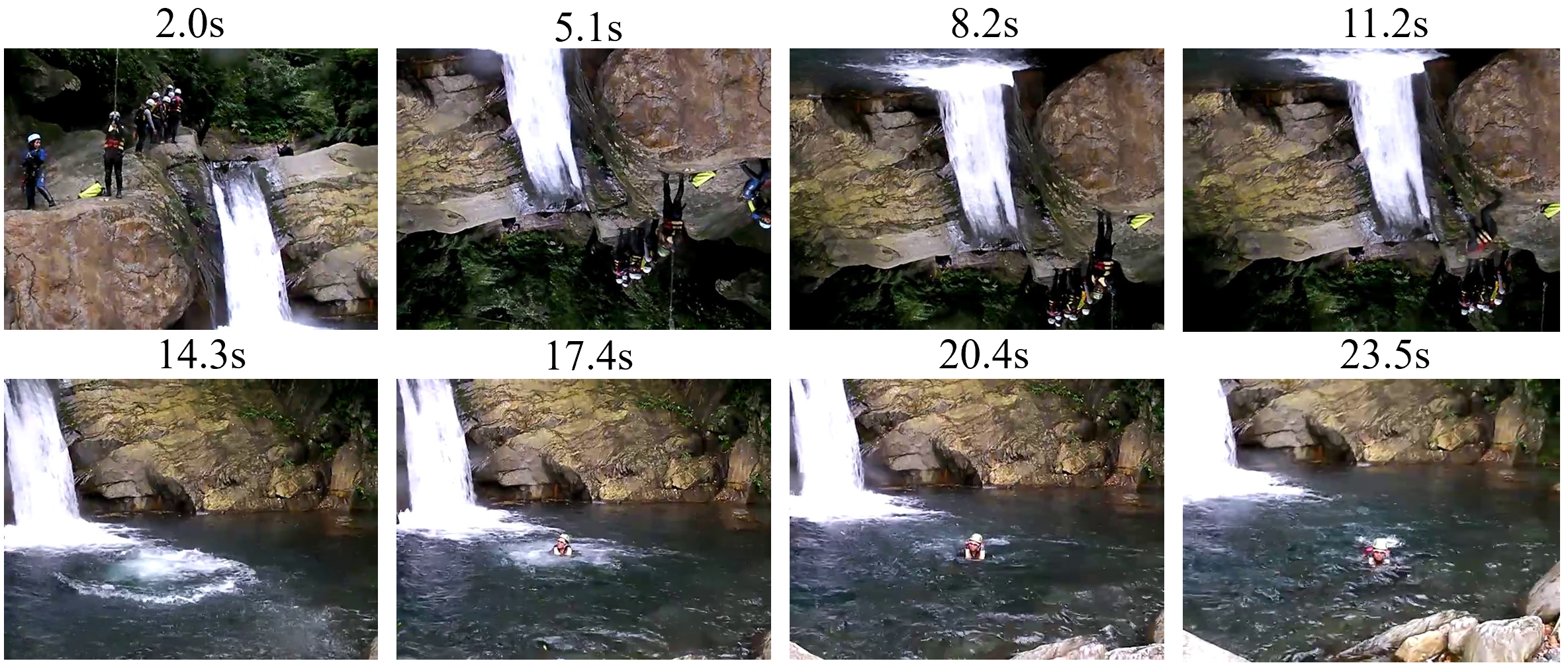}
    \caption{\textbf{An example of Rotate.} The ground truth is 5.1s--11.7s.}
    \label{fig:demo_rotate}
\end{figure*}

\begin{figure*}[t]
    \centering
    \includegraphics[width=\textwidth]{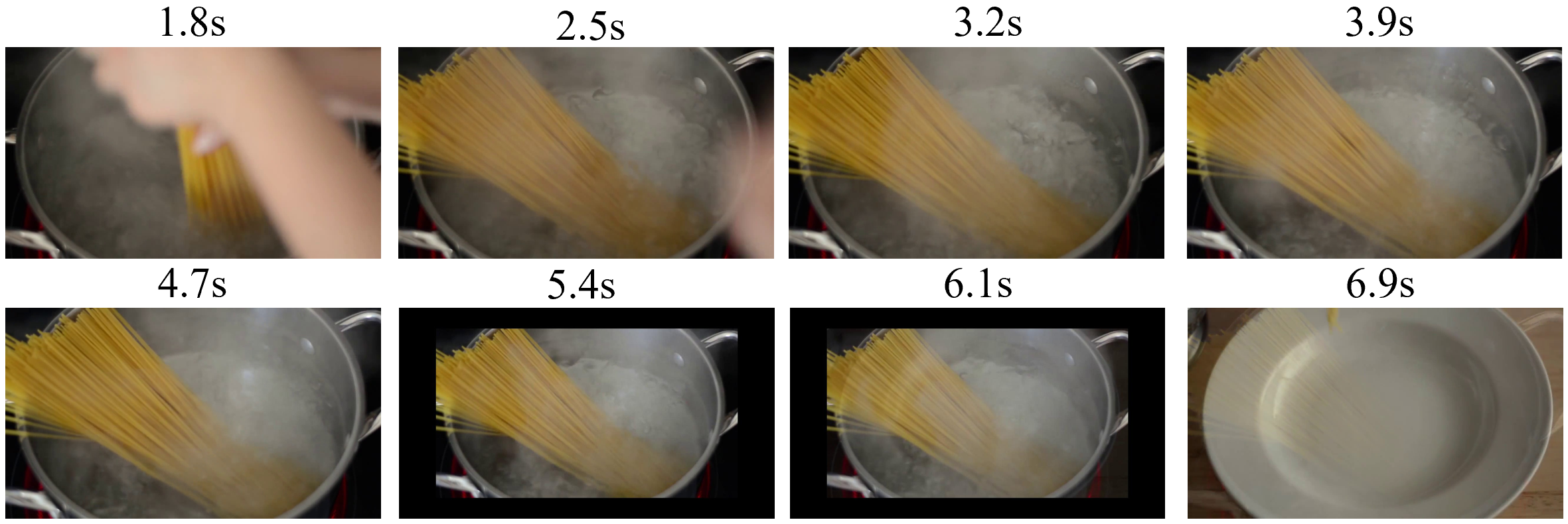}
    \caption{\textbf{An example of ZoomOut.} The ground truth is 5.4s--6.9s.}
    \label{fig:demo_zoomout}
\end{figure*}

\begin{figure*}[t]
    \centering
    \includegraphics[width=\textwidth]{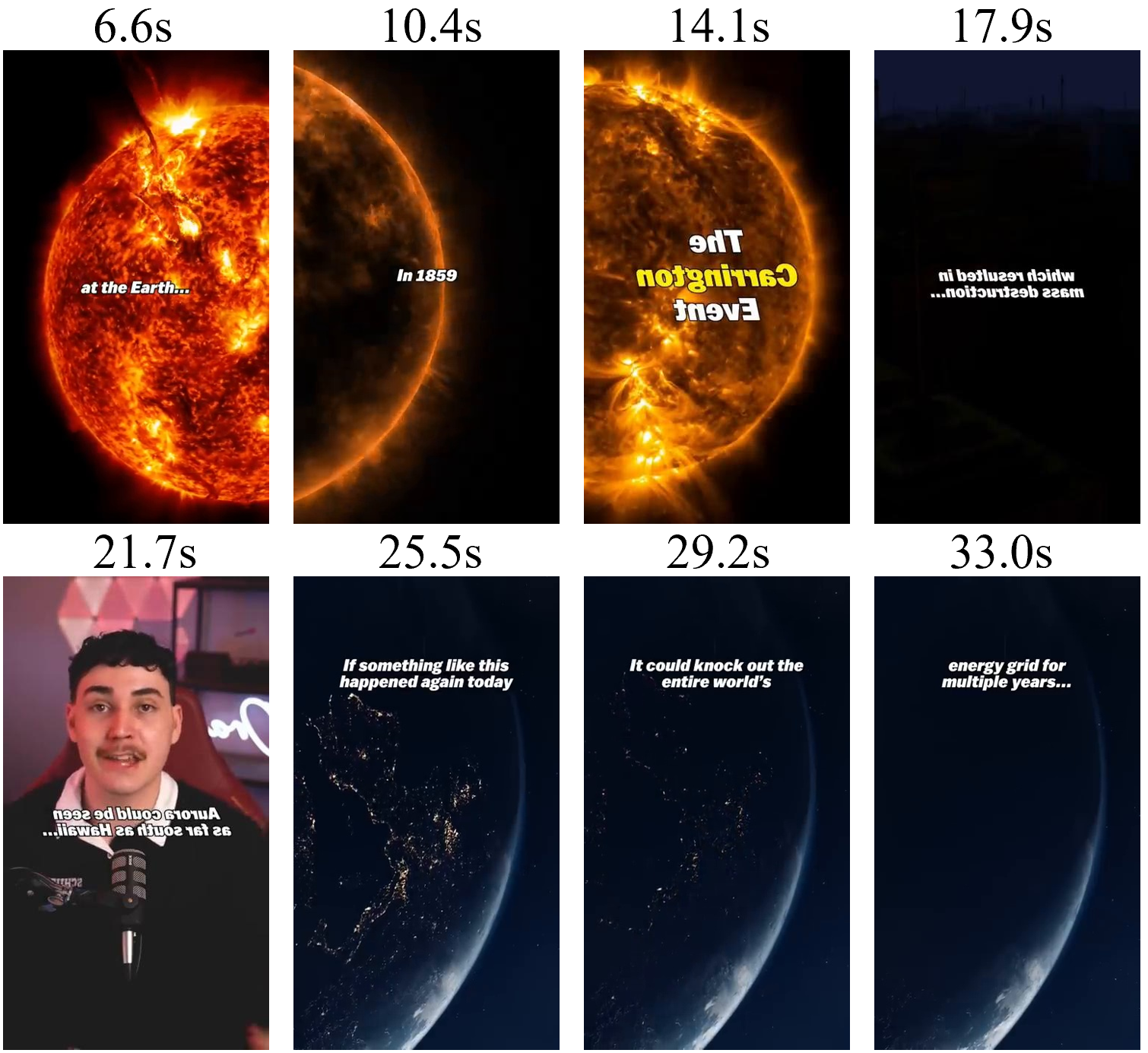}
    \caption{\textbf{An example of Mirror.} The ground truth is 14.1s--22.6s.}
    \label{fig:demo_mirror}
\end{figure*}

Figures~\ref{fig:demo_channel} through \ref{fig:demo_mirror} illustrate several concrete examples of the Anomaly Grounding task, corresponding to four different perturbation types. For clarity, only a subset of key frames from each video is displayed. The model's objective is to predict the temporal range of the introduced anomaly based on the visual evidence.

\subsection{Object Counting}
\label{subsec:example_object_counting}

\begin{figure}[h!]
    \centering
    \includegraphics[width=\columnwidth]{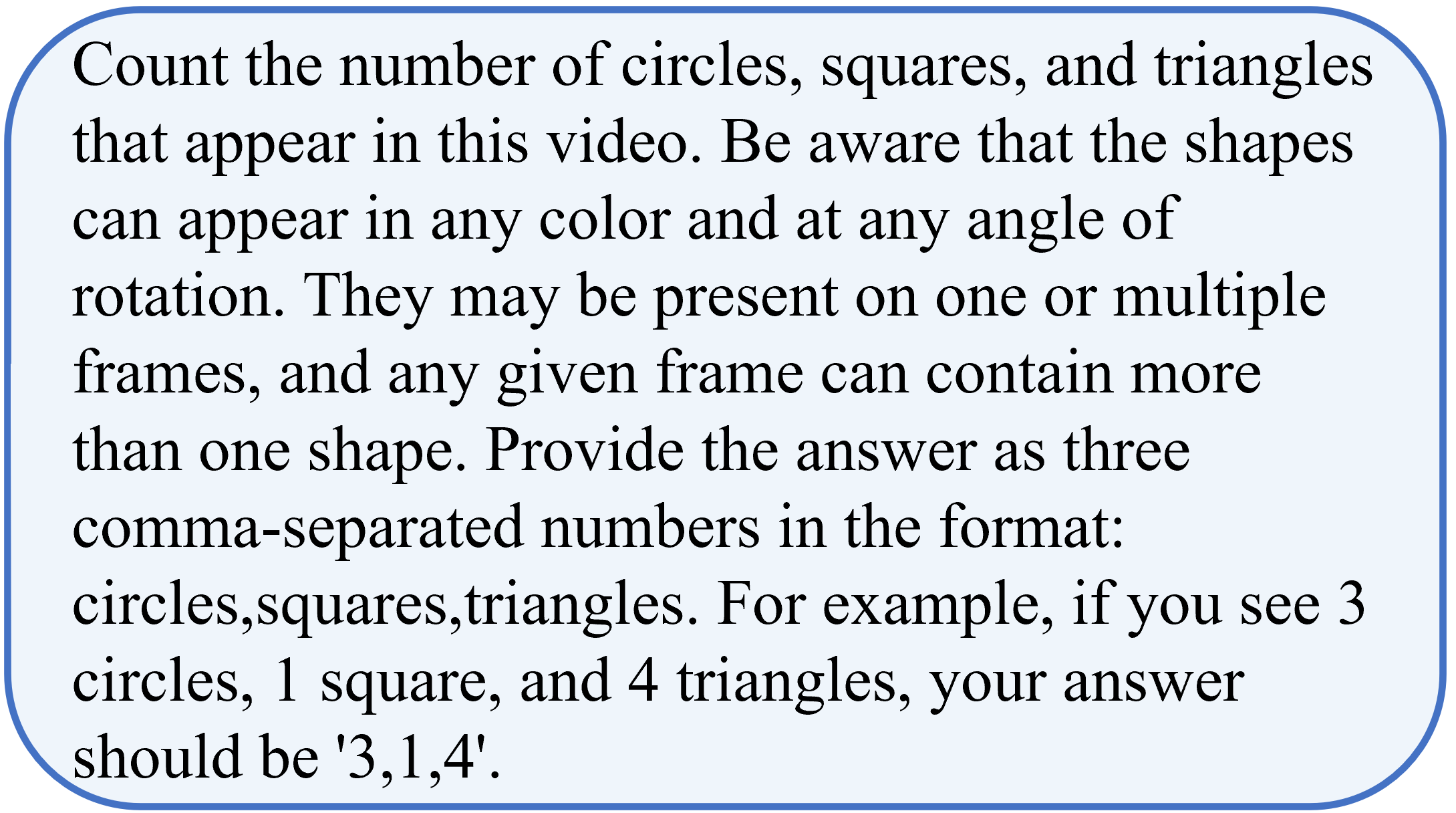} 
    \caption{\textbf{Prompt template for Object Counting.}}
    \label{fig:task_counting}
\end{figure}

\begin{figure*}[t]
    \centering
    \includegraphics[width=\textwidth]{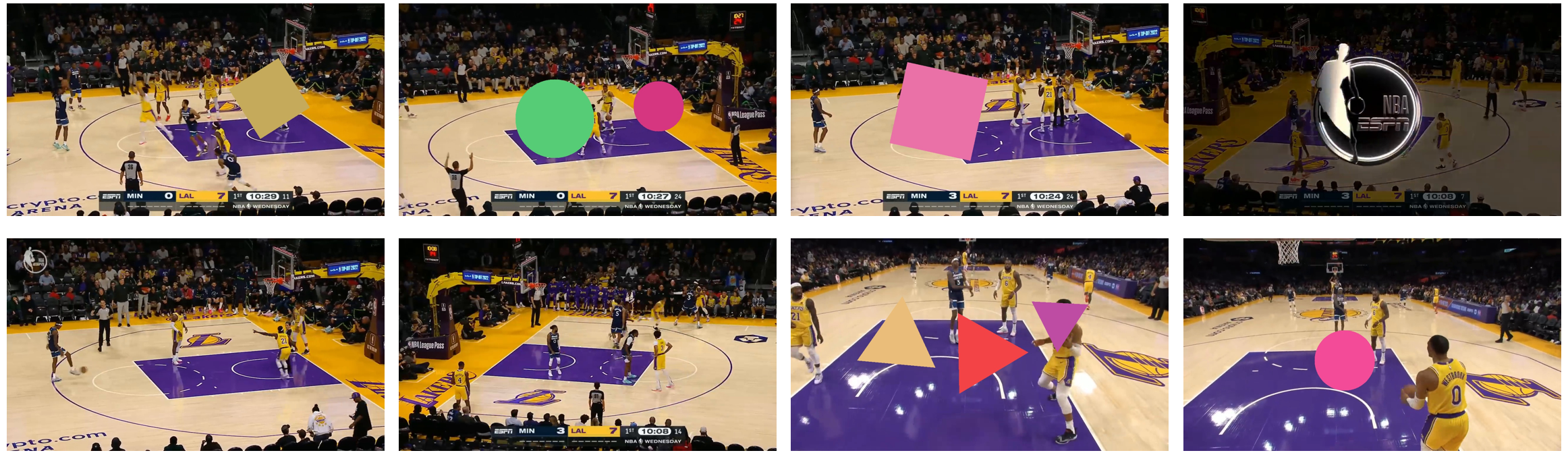}
    \caption{\textbf{An example of Object Counting.} The ground truth (circles, squares, and triangles) is 3,2,3.}
    \label{fig:demo_counting}
\end{figure*}

\begin{figure*}[t]
    \centering
    \includegraphics[width=\textwidth]{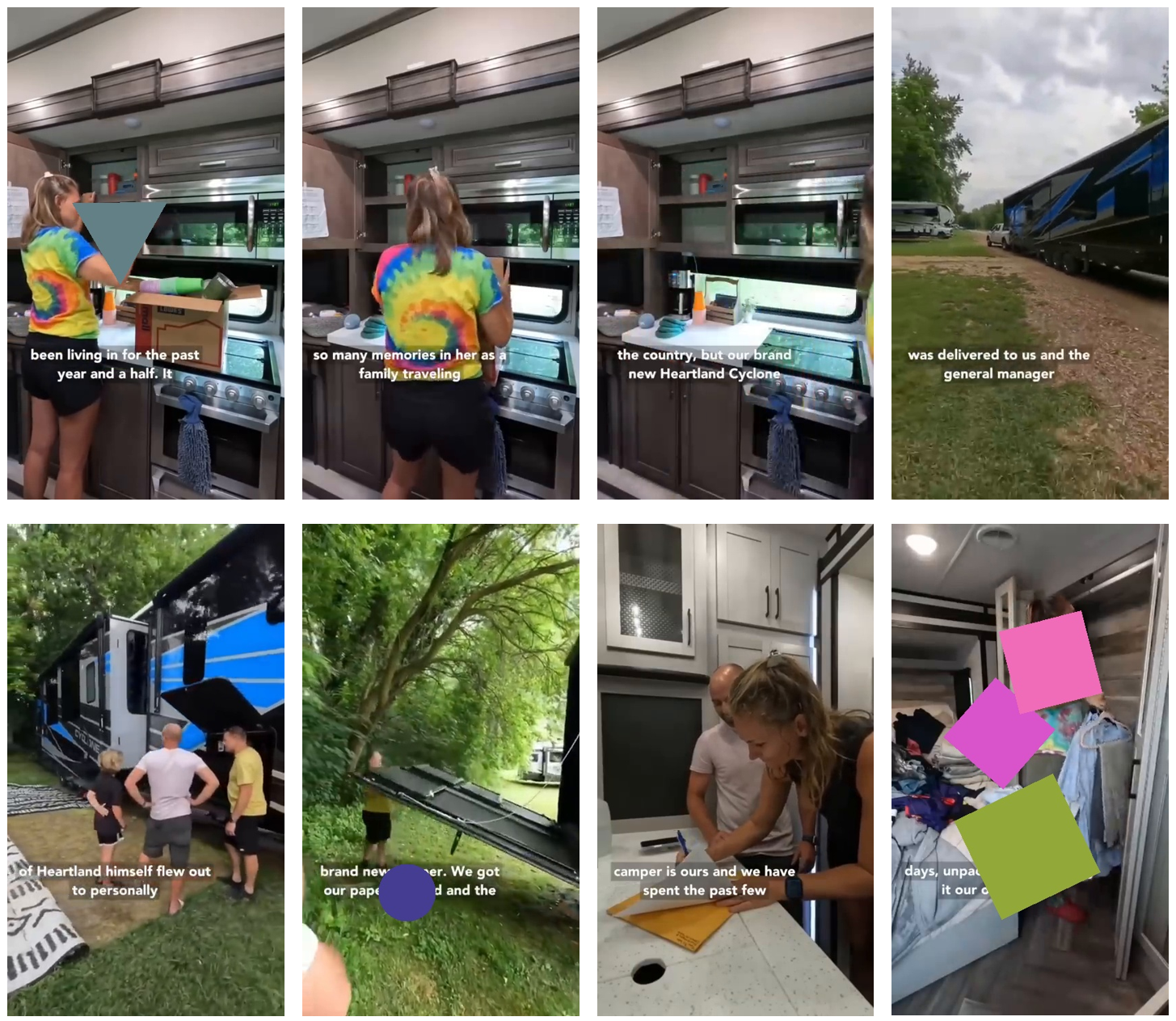}
    \caption{\textbf{An example of Object Counting.} The ground truth (circles, squares, and triangles) is 1,3,1.}
    \label{fig:demo_counting2}
\end{figure*}

Figure~\ref{fig:task_counting} illustrates the prompt template used for the Object Counting task. Concrete visual examples of this task are provided in Figure~\ref{fig:demo_counting} and Figure~\ref{fig:demo_counting2}.

\subsection{Temporal Jigsaw}
\label{subsec:example_temporal_jigsaw}
\begin{figure}[h!]
    \centering
    \includegraphics[width=\columnwidth]{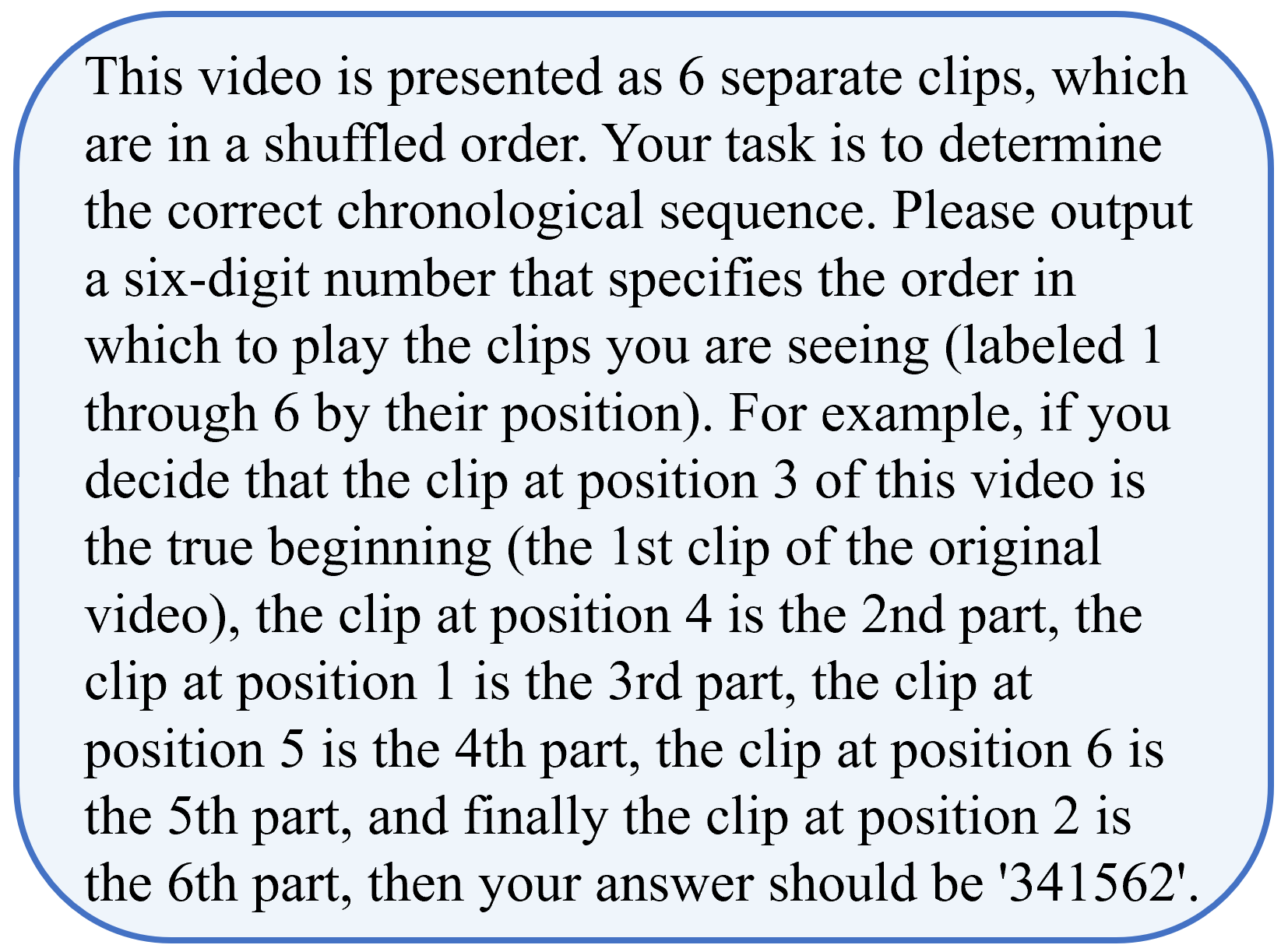} 
    \caption{\textbf{Prompt template for Temporal Jigsaw.}}
    \label{fig:task_jigsaw_prompt}
\end{figure}

\begin{figure*}[t]
    \centering
    \includegraphics[width=\textwidth]{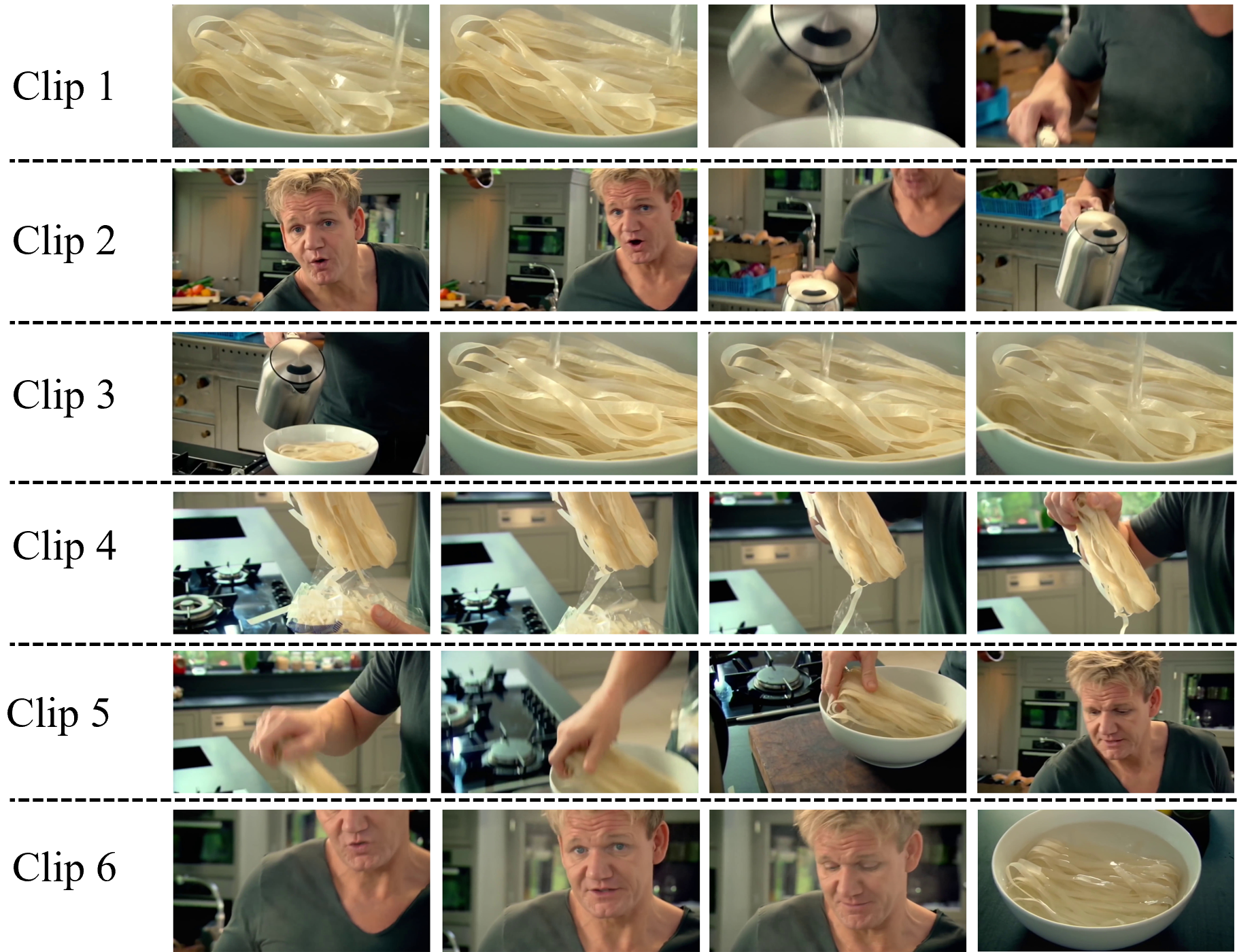}
    \caption{\textbf{An example of Temporal Jigsaw.} The ground truth is 452316. The corresponding unshuffled video is shown in Figure~\ref{fig:demo_jigsaw_original}.}
    \label{fig:demo_jigsaw}
\end{figure*}

\begin{figure*}[t]
    \centering
    \includegraphics[width=\textwidth]{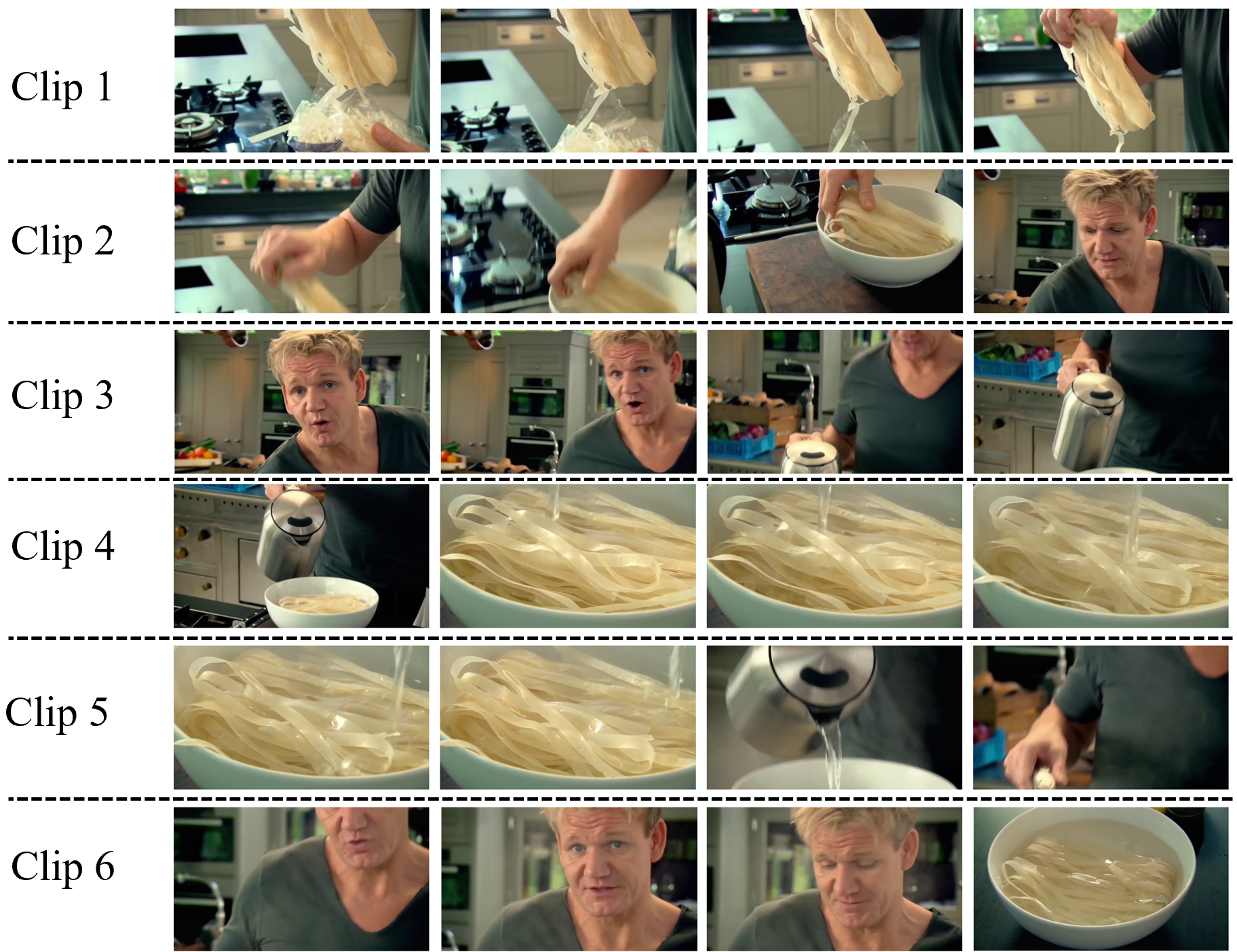}
    \caption{\textbf{The original video corresponding to the example in Figure~\ref{fig:demo_jigsaw}.}}
    \label{fig:demo_jigsaw_original}
\end{figure*}

Figure~\ref{fig:task_jigsaw_prompt} shows the prompt template for the Temporal Jigsaw task. Figure~\ref{fig:demo_jigsaw} provides a concrete visual example of the shuffled video sequence that is presented to the model. For a clearer understanding of the task and to provide a direct comparison, the corresponding original video with the clips in their correct temporal order is also shown in Figure~\ref{fig:demo_jigsaw_original}.

\subsection{Exploration of Alternative Pretext Tasks}
\label{subsec:alternative_tasks}

In addition to the three pretext tasks detailed in the main paper, we also investigated other self-supervised learning paradigms. Our exploration included generative modeling approaches, such as masked~\cite{tong2022videomae,chen2025versavid} frame reconstruction and autoregressive~\cite{wang2025fostering,yang2025cambrian} next frame prediction. Furthermore, we experimented with a task focused on direct temporal speed prediction~\cite{wang2020self}.

However, our preliminary experiments indicated that these alternative tasks did not yield significant or consistent performance improvements on our downstream evaluation benchmarks. This suggests that while these methods are powerful, their objectives may not be as directly aligned with cultivating the high level perceptual and reasoning skills targeted by our final task selection. The discovery of an even broader range of effective self-supervised tasks for enhancing MLLMs remains a promising direction for future work.

\section{Limitations and Future Work}
\label{sec:limitations}

While our work demonstrates the significant potential of VideoSSR, we also recognize several limitations that present clear opportunities for future research.

First, our experiments were primarily conducted using a low number of input frames for both training and evaluation. This decision was driven by considerations of computational efficiency, allowing for both rapid iteration on more pretext tasks and comprehensive coverage of evaluation benchmarks. However, this approach may limit the model's scalability to long videos. A key direction for future work is to scale the VideoSSR framework to handle higher frame rates and longer video inputs. This will be crucial for enhancing the model's capabilities on complex, long-form content where dense temporal information is paramount.

Second, our framework relies on only three pretext tasks. While effective, this approach overlooks both the potential of a broader range of self-supervised objectives and the possible synergies that could be unlocked with more sophisticated mixing strategies. Future work could therefore explore a richer suite of pretext tasks and investigate advanced mixing techniques like curriculum learning or adaptive task weighting to further enhance model generalization.

\clearpage

